\documentclass[10pt,twocolumn,letterpaper]{article}

\usepackage{iccv}              

\definecolor{iccvblue}{rgb}{0.21,0.49,0.74}
\definecolor{darkgreen}{RGB}{0,128,0} 
\usepackage[pagebackref,breaklinks,colorlinks,allcolors=iccvblue]{hyperref}
\usepackage{colortbl}
\usepackage{multirow}
\usepackage{pifont}

\def\paperID{14520} 
\def\confName{ICCV}
\def\confYear{2025}

\title{DTFSal: Audio-Visual Dynamic Token Fusion for Video Saliency Prediction}

\author{
Kiana Hooshanfar$^{1}$, 
Alireza Hosseini$^{1}$, 
Ahmad Kalhor$^{1}$,
Babak Nadjar Araabi$^{1}$ \\
$^1$University of Tehran \\
{\tt\small \{k.hooshanfar, arhosseini77, Kalhor, Araabi\}@ut.ac.ir} \\
\normalsize \url{https://k-hooshanfar.github.io/dtfsal/}
}

\begin{document}
\maketitle
\begin{abstract}
Audio-visual saliency prediction aims to mimic human visual attention by identifying salient regions in videos through the integration of both visual and auditory information. Although visual-only approaches have significantly advanced, effectively incorporating auditory cues remains challenging due to complex spatio–temporal interactions and high computational demands. To address these challenges, we propose \textbf{D}ynamic \textbf{T}oken \textbf{F}usion \textbf{Sal}iency (DFTSal), a novel audio-visual saliency prediction framework designed to balance accuracy with computational efficiency. Our approach features a multi-scale visual encoder equipped with two novel modules: the Learnable Token Enhancement Block (LTEB), which adaptively weights tokens to emphasize crucial saliency cues, and the Dynamic Learnable Token Fusion Block (DLTFB), which employs a shifting operation to reorganize and merge features, effectively capturing long-range dependencies and detailed spatial information. In parallel, an audio branch processes raw audio signals to extract meaningful auditory features. Both visual and audio features are integrated using our Adaptive Multimodal Fusion Block (AMFB), which employs local, global, and adaptive fusion streams for precise cross-modal fusion. The resulting fused features are processed by a hierarchical multi-decoder structure, producing accurate saliency maps. Extensive evaluations on six audio-visual benchmarks demonstrate that DFTSal achieves SOTA performance while maintaining computational efficiency.
\end{abstract}    
\vspace{-12pt}
\section{Introduction}
\label{sec:intro}
\vspace{-5pt}
Human vision naturally identifies salient regions in dynamic scenes, guiding attention and enabling efficient information processing \cite{koch2006much}. In computer vision, saliency prediction models attempt to replicate this human capability by generating saliency maps that highlight where observers are most likely to focus~\cite{sun2017}. These models are essential in a wide range of applications, including marketing~\cite{hosseini2024brand, jiang2023ueyes}, multimedia~\cite{mishra2021multi}, object detection~\cite{bartolo2024correlation, zhou2021saliency}, image/video compression~\cite{gungordu2024saliency}, user-interface design~\cite{jiang2023ueyes}, and cognitive research~\cite{das2024shifting}. Although significant progress has been made for static images~\cite{hosseini2024sum, aydemir2023tempsal}, predicting saliency in videos remains challenging due to the need to capture multi-scale spatial-temporal cues and handle complex motion.

Recent video saliency methods have employed deep learning architectures such as CNN-LSTM frameworks and transformer-based networks to improve performance \cite{Wang2018revisiting, Wu2020, jain2021vinet, Chang2023, Ma2022, zhou2023transformer}. At the same time, incorporating auditory information has proven beneficial for capturing salient events in complex scenarios \cite{zhu2024does}, as demonstrated by DAVE \cite{tavakoli2019dave} and STAViS \cite{tsiami2020stavis}. Meanwhile, models such as CASP-Net \cite{xiong2023casp}, DiffSal \cite{xiong2024diffsal} and RAVF \cite{yu2024relevance} integrate audio-visual cues through various cross-modal fusion strategies, but often suffer from large-scale networks, complex attention mechanisms, or incomplete spatio–temporal modeling~\cite{hu2023tinyhd}. These limitations underscore a persistent challenge, achieving high predictive accuracy while maintaining computational efficiency and robust cross-modal alignment, particularly when dealing with dynamic audio-visual data in resource-constrained scenarios.

To address these challenges, we propose a novel audio-visual saliency framework, \underline{D}ynamic \underline{T}oken \underline{F}usion \underline{Sal}iency Prediction (DTFSAL). Our approach leverages a multi-scale visual encoder to capture detailed spatial-temporal features from videos, while a dedicated audio branch processes audio signals. We then introduce three key modules to refine and fuse these representations:

\noindent \textbf{Learnable Token Enhancement Block (LTEB)}: Adaptively emphasizes crucial spatio-temporal cues by weighting salient tokens and suppressing less relevant features.

\noindent \textbf{Dynamic Learnable Token Fusion Block (DLTFB)}: Employs a shifting mechanism that reorganizes and merges tokens, improving long-range dependencies and preserving important spatial details.

\noindent \textbf{Adaptive Multimodal Fusion Block (AMFB)}: Integrates audio and visual features via three complementary streams (local, global, and adaptive), enabling precise cross-modal alignment.

Finally, a hierarchical multi-decoder generates multi-scale saliency maps, which are combined through a final 3D convolution to yield a coherent saliency prediction.
In summary, our main \textbf{contributions} are:
\begin{itemize}
    \item We propose DTFSal, a novel audio-visual saliency model that efficiently fuses multi-scale video and audio features for improved performance.
    \item We introduce three specialized modules (LTEB, DLTFB, and AMFB) that jointly refine spatio-temporal features, capture long-range dependencies, and adaptively fuse cross-modal information.
    \item We demonstrate state-of-the-art (SOTA) results on challenging audio-visual saliency benchmarks while preserving computational efficiency, showing that our approach achieves a balance between accuracy and resource demands.
\end{itemize}

\begin{figure*}[t]
    \centering
    \includegraphics[width=\textwidth]{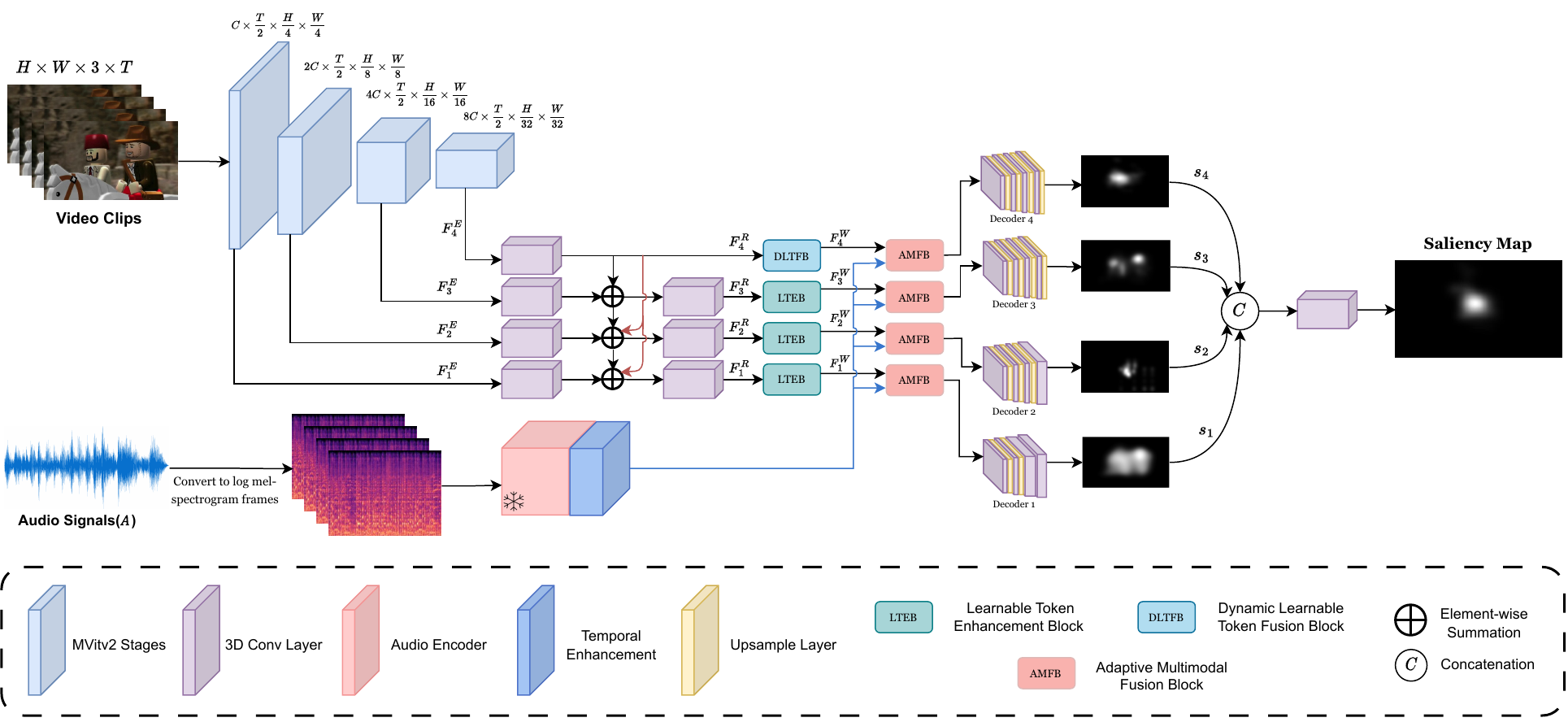}
    \caption{Overview of our DTFSal model, which integrates a multi-scale encoder, a hierarchical multi-decoder, LTEB, DLTFB, and AMFB for efficient and accurate audio-visual saliency prediction.}
    \label{fig:dtfsal_Network}
    \vspace{-1em}
\end{figure*}

\vspace{-5pt}
\section{Related Work} \label{sec:related}

\textbf{Visual Saliency Prediction:} Early video saliency prediction methods relied on motion-based models and manual feature extraction but struggled to capture spatio-temporal relationships, limiting their effectiveness in tracking object motions over time. With the rise of deep learning, these traditional approaches have been largely replaced by more advanced models capable of learning complex spatial and temporal patterns directly from video data. Early approaches extended image-based models with temporal modules, such as ACLNet~\cite{wang2019revisiting}, an attentive CNN-LSTM that extracts frame-wise features and processes them with a ConvLSTM enhanced by a supervised spatial attention module trained on large static saliency datasets. Other comparable models utilized two-stream architectures that separately processed appearance and motion while incorporating an objectness stream to capture salient objects, enhancing performance at the cost of increased model size \cite{hu2023tinyhd, kocak2021gated, zhang2023multi}. These CNN+RNN and multi-stream networks showcased the effectiveness of deep spatio-temporal feature fusion for saliency prediction but often struggled to capture long-range dependencies in complex scenes \cite{min2019tased, zhang2018video}.

More recent methods \cite{xiong2023casp, zhou2023transformer, wu2023gfnet, moradi2024transformer} have increasingly adopted Transformer-based models to overcome these limitations. Transformers, using self-attention mechanisms, effectively capture long-range and multi-scale temporal dependencies in video data. For instance, Wang \emph{et al.}\cite{xiong2023casp} integrate self-attention into a 3D CNN backbone to model global spatial-temporal contexts, while other works, such as TMFI-Net \cite{zhou2023transformer}, Combine multi-scale temporal features through Transformer-based feature integration to achieve SOTA accuracy, though at the cost of large model sizes \cite{tsiami2020stavis}. While Transformer-enhanced methods deliver remarkable performance, they highlight a trade-off between capturing global context and ensuring computational efficiency. UNISAL and TinyHD offers a more lightweight alternative; however, it falls short in accuracy and robustness compared to newer architectures that achieve a more optimized balance \cite{droste2020unified, hu2023tinyhd, yu2024relevance}.

\noindent \textbf{Audio-Visual Saliency Prediction:}
Incorporating audio cues into video saliency models has become a growing focus due to the significant influence of sound on visual attention \cite{zhu2024discrete, yu2024relevance, xie2024audio}. The DAVE model \cite{tavakoli2019dave} illustrates this approach, introducing the Audio-Visual Eye-tracking dataset (AVE) and a simple encoder-decoder that integrates audio and visual features to enhance saliency prediction \cite{tavakoli2019dave}. Subsequently, STAViS, a spatio-temporal audio-visual saliency network, demonstrated that synchronized audio, such as speech and noise events, enhances saliency prediction in dynamic scenes \cite{tsiami2020stavis}. More specialized models \cite{liu2020learning, qiao2021joint} have been developed to focus on multi-speaker conversation videos, predicting saliency and localizing sound sources. Advancements like CASP-Net \cite{xiong2023casp} and DiffSal \cite{xiong2024diffsal} further improve performance through enhanced fusion strategies and increased model complexities, scaling up to 76M parameters. A newer methodology, DAVS \cite{zhu2024discrete}, leverages Implicit Neural Networks~\cite{sitzmann2020implicit} for audio-visual saliency prediction, effectively mapping spatio-temporal coordinates to saliency values.
In contrast to previous approaches that often compromise either global context or efficiency, our proposed DFTSal framework dynamically fuses visual and auditory cues to capture long-range dependencies without incurring heavy computational costs. By effectively integrating audio and visual cues in dynamic scenes, our method bridges the gap between accuracy and efficiency.

\vspace{-7.5pt}
\section{Proposed Method} \label{sec:method}
\vspace{-3pt}
DTFSal, as shown in~\autoref{fig:dtfsal_Network}, leverages a pre-trained encoder to extract multi-scale features from video clips. Two novel modules further enhance these features: LTEB adaptively highlights key cues, while DLTFB reorganizes and merges spatial and temporal tokens. A dedicated audio branch extracts features from raw audio and aligns them with visual cues. Finally, AMFB fuses these cues through local, global, and adaptive streams, feeding into a hierarchical multi-decoder that integrates multi-scale information. A final 3D convolution then merges the decoder outputs to produce a confident saliency map.

\begin{figure}[b]
    \centering
    \resizebox{0.9\columnwidth}{!}{%
        \includegraphics{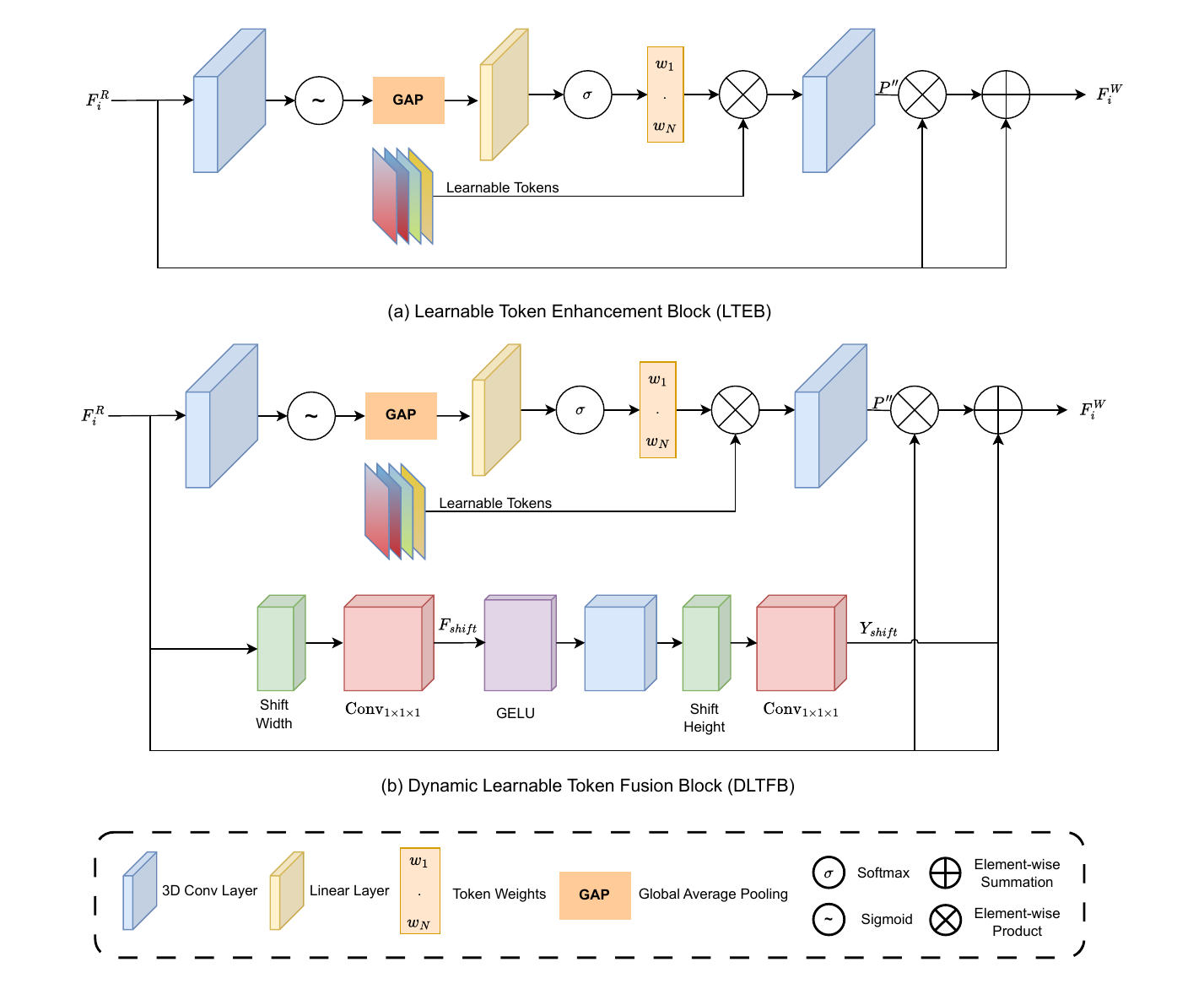}
    }
    \caption{Overview of LTEB and DLTFB: (a) LTEB emphasizes salient cues by adaptively weighting tokens. (b) DLTFB enhances spatial feature integration by shifting and fusing tokens.}
    \label{fig:LTEB_DLTFB}
\end{figure}

\vspace{-3pt}

\subsection{Two-Stream Encoders}

\noindent \textbf{Visual Encoder:} Our visual encoder uses MViTv2 architecture~\cite{li2022mvitv2}, pre-trained on Kinetics-400~\cite{carreira2017quo}, which employs a multiscale pyramid approach as in~\cite{fan2021multiscale}. The encoder is pre-trained on large-scale human action recognition dataset, enhancing feature representation and spatio-temporal pattern capture for improved performance  \cite{pan2017salgan,min2019tased}. A video clip of size \(T \times H \times W \times 3\) is first split by a patch partition layer into non-overlapping tokens of size \(T/2 \times H/4 \times W/4\), with each token represented by a 96-dimensional feature vector. These tokens then pass through a series of stages that gradually reduce the spatial resolution while increasing the channel capacity, forming a multilevel feature pyramid:

\vspace{-9pt}

\begin{equation}
\{\mathbf{F}^E_i \in \mathbb{R}^{(2^{i-1} \times C) \times \frac{T}{2} \times \frac{H}{2^{i+1}} \times \frac{W}{2^{i+1}}}\}_{i=1}^4,
\end{equation}

\noindent where \(C=96\). This hierarchical design efficiently captures both local details and long-range dependencies.

\vspace{3pt}

\noindent \textbf{Audio Encoder:} To synchronize audio with visual features, the raw audio \((A)\) is first converted into a Log-Mel Spectrum via Short-time Fourier transform and segmented into \(T_a\) parts of size \(H_a \times W_a \times 1\) using an 11-ms overlapping window. Each audio segment \(f_{a,i}\) (for \(i=1,\dots,T_a\)) is processed by a fully convolutional 2D VGGish network~\cite{hershey2017cnn}, pre-trained on AudioSet~\cite{gemmeke2017audio}, to yield features of shape \(\mathbb{R}^{h_a \times w_a \times C_a}\). A temporal enhancement module, comprising a patch embedding layer followed by a Transformer layer, is then applied to ensure robust temporal alignment~\cite{xiong2024diffsal}.

\vspace{-5pt}

\subsection{Feature Enhancement}

\vspace{-5pt}

A top–down pathway with lateral connections enriches the semantic and spatio–temporal information of the extracted features. This process is performed in three steps. In the first step, 3D convolutional layers adjust the temporal and channel dimensions of the features, yielding refined multi-scale representations:
\vspace{-3.3pt}
\begin{equation} 
F_i \in \mathbb{R}^{C^* \times \frac{T}{2} \times \frac{H}{2^{i+1}} \times \frac{W}{2^{i+1}}}, \quad (C^* = 192, ; i=1,2,3,4). 
\end{equation}

In the second step, high-level features are progressively injected into lower levels via a top–down pathway with lateral connections. Finally, in the third step, the fused features are further processed by additional 3D convolutional layers to produce the refined multi-scale features \(F_i^R\). The entire process is summarized as follows:
\vspace{-3pt}
\begin{equation}
\begin{split}
    F_4^R &= \text{Conv}(F_4^E), \\
    F_3 &= \text{Conv}(F_3^E) \oplus U_{p\times2}(F_4^R), \\
    F_2 &= \text{Conv}(F_2^E) \oplus U_{p\times2}(F_3) \oplus U_{p\times4}(F_4^R), \\
    F_1 &= \text{Conv}(F_1^E) \oplus U_{p\times2}(F_2) \oplus U_{p\times8}(F_4^R), \\
    F_i^R &= \text{Conv}(F_i), \quad i=3,2,1.
\end{split}
\label{eq:feature_enhancement}
\end{equation}
\noindent In these equations, \(\text{Conv}(\cdot)\) dsenotes a 3D convolution, \(U_{p \times i}(\cdot)\) (\(i=2,4,8\)) represents trilinear upsampling, and \(\oplus\) is the element-wise addition. This process produces robust semantic and multi-scale features that capture precise information across different spatial and temporal locations.

\subsubsection{Learnable Token Enhancement Block (LTEB)}
To further refine the spatio–temporal features, we introduce LTEB, which selectively emphasizes salient cues while suppressing redundant information. LTEB first applies a 3D convolution followed by a sigmoid activation that dynamically weights the input features across temporal, spatial, and channel dimensions. This mechanism works as follows:
\vspace{-5.5pt}
\begin{equation}
	G = \text{Sigmoid}\left(\text{Conv}_{1 \times 3 \times 3}(F^R)\right),
	\label{eq:lteb_gate}
\end{equation}
After that, a global embedding (\(\mathbf{E}(t)\)) is computed by spatially averaging \(G\).
We obtain the adaptive token weights \(\{w_i\}_{i=1}^{N}\) by applying a softmax on a linear transformation of \(\mathbf{E}(t)\):
\vspace{-5.5pt}
\begin{equation}
	\{w_i\}_{i=1}^{N} = \text{Softmax}\Bigl(\text{Linear}\bigl(\mathbf{E}(t)\bigr)\Bigr).
\end{equation}
To adaptively adjust the contribution of each token based on its significance, we apply element-wise multiplication between these weights and the generated learnable tokens \(\{\mathbf{P}_i\}_{i=1}^{N}\), producing refined representations, followed by interpolation and a 3D convolution to yield:

\vspace{-5.5pt}
\begin{equation}
	\mathbf{K} = \sum_{i=1}^{N} w_i \cdot \mathbf{P}_i\,(H, W),
	\label{eq:lteb_token1}
\end{equation}

\vspace{-5.5pt}
\begin{equation}
	\mathbf{P}'' = \text{Conv}_{1 \times 3 \times 3}\Bigl( \text{Interpolate}\bigl(\mathbf{K}\bigr) \Bigr).
	\label{eq:lteb_token2}
\end{equation}

Here, the weights serve as attention scores derived from the input features, dynamically modulating each token’s influence. The learnable tokens collectively form a unified embedding space, promoting correlated knowledge sharing throughout the model. Finally, a residual connection fuses the original features with the adaptively weighted tokens:
\vspace{-5.5pt}
\begin{equation}
	\mathbf{F^w} = \mathbf{F} \cdot \mathbf{P}'' + \mathbf{F}.
	\label{eq:lteb_final}
\end{equation}

This mechanism, as shown in ~\autoref{fig:LTEB_DLTFB} (a), dynamically highlights the most informative spatio–temporal cues while preserving structural consistency. In essence, LTEB recalibrates the feature map by amplifying salient regions and suppressing noise through adaptive token weighting. This refinement leads to a more discriminative and robust representation, ensuring that subsequent modules operate on features that are both precise and computationally efficient.

\subsubsection{Dynamic Learnable Token Fusion Block (DLTFB)}
Building upon LTEB, DLTFB further refines the spatial distribution of features by employing a dynamic shifting mechanism. Inspired by the Swin Transformer~\cite{liu2021swin} and axial-attention~\cite{wang2020axial}, the shifting operations along the width and height dimensions are introduced to enhance the spatial arrangement of the features. In DLTFB, adaptive token computation is performed independently while simultaneously applying spatial shifting operations to rearrange the feature maps. First, the input features are shifted along the width dimension (\(F_{\text{shift}}\)) and then processed by a convolution followed by a 3D convolution with GELU activation:
\vspace{-5pt}
\begin{equation}
	Y = \operatorname{Conv3d}\Bigl(\operatorname{GELU}\bigl(\text{Conv}_{1 \times 1 \times 1}(\mathbf{F}_{\text{shift}})\bigr)\Bigr).
	\label{eq:dltf_conv}
\end{equation}
A subsequent shift along the height dimension, \(Y_{\text{shift}}\), is followed by layer normalization (LN) and a convolution, producing the refined features:

\vspace{-5pt}
\begin{equation}
	\mathbf{F}^{Sh} = F + \text{Conv}_{1 \times 1 \times 1}\Bigl( \operatorname{LN}\bigl(Y_{\text{shift}}\bigr) \Bigr).
	\label{eq:dltf_norm}
\end{equation}
Finally, the output of DLTFB fuses the weighted original features with the spatially refined information:
\vspace{-5pt}
\begin{equation}
	\mathbf{F}^{w} = F \cdot \mathbf{P}'' + \mathbf{F}^{Sh}.
	\label{eq:dltf_final}
\end{equation}

\noindent This block, as illustrated in~\autoref{fig:LTEB_DLTFB} (b), enhances the model's capability to capture complex local patterns and long-range dependencies by dynamically rearranging spatial features. In simple terms, DLTFB reorders and mixes parts of the feature map so that information from salient regions is better integrated. This leads to a richer and more context-aware representation, which is crucial for accurate saliency prediction in challenging and dynamic scenes. Moreover, by fusing these refined features with the original ones, DLTFB ensures that both detailed and overall structural information is preserved, resulting in a more robust and efficient model.

\begin{figure*}[t]
    \centering
    \includegraphics[width=\textwidth]{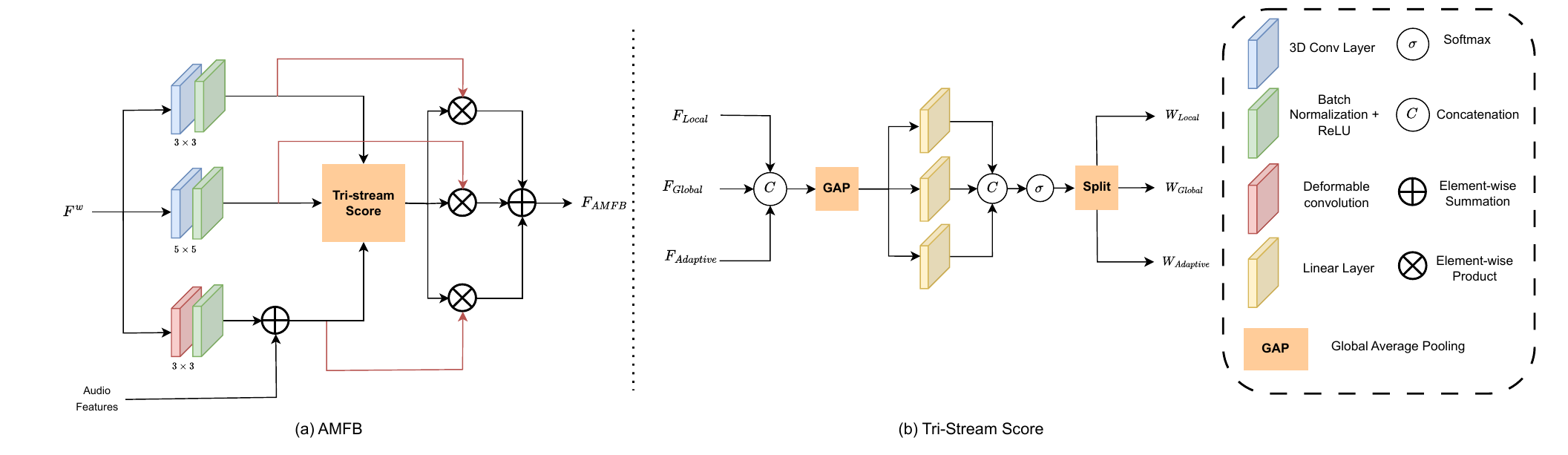}
    \caption{Overview of AMFB. (a) AMFB integrates local, global, and adaptive streams to fuse audiovisual cues. (b) Tri-Stream Score module computes attention weights via GAP, a linear layer, and sigmoid activation.}
    \label{fig:AMFB}
\end{figure*}

\subsection{Adaptive Multimodal Fusion Block (AMFB)}
AMFB, as shown in~\autoref{fig:AMFB} (a), effectively integrates audio and visual information by capturing complementary cues at multiple scales. In our design, the audio-visual inputs are processed through three independent streams, each extracting distinct signal aspects:

\noindent \textbf{Local Stream:} A 3D convolution with a \(3 \times 3\) kernel (accompanied by Batch Normalization and ReLU) extracts fine-grained, spatially localized features, which are crucial for detecting precise details and identifying key visual cues.

\noindent \textbf{Global Stream:} A 3D convolution with a 
 \(5 \times 5\) kernel aggregates a broader, scene-level context. This stream captures overall cues and the general scene structure, enabling the model to adjust saliency based on the larger context.

\noindent \textbf{Adaptive Stream:} This branch leverages a \(3 \times 3\) deformable convolution~\cite{dai2017deformable} with learned offsets to adjust its receptive field, effectively handling objects at different scales and capturing fine details, as noted in~\cite{jarimijafarbigloo2024reducing}. Additionally, audio features extracted via the VGGish encoder and integrated for cross-modal alignment.

The outputs of the three streams—represented as \(F_{\text{Local}}\), \(F_{\text{Global}}\), and \(F_{\text{adaptive}}\)—are concatenated along the channel dimension referred to as \(F_{\text{fused}}\):
\vspace{-3pt}
\begin{equation}
F_{\text{fused}} = \bigl[F_{\text{Local}}: F_{\text{Global}}: F_{\text{Adaptive}}\bigr].
\label{eq:amfb_fuse}
\end{equation}
An adaptive attention mechanism is then applied, referred to as the \emph{Tri-Stream Score module}: global average pooling (GAP) is applied to \(F_{\text{fused}}\), which is then passed through a linear layer followed by a sigmoid activation to compute the Tri-Stream Scores (as illustrated in~\autoref{fig:AMFB} (b)):
\vspace{-3pt}
\begin{equation}
\mathbf{W} = \text{Sigmoid}\Bigl(\text{Linear}\bigl(\text{GAP}(F_{\text{fused}})\bigr)\Bigr),
\label{eq:amfb_scores}
\end{equation}
\noindent where \(\mathbf{W} = [W_{\text{Local}},\, W_{\text{Global}},\, W_{\text{Adaptive}}]\). The final fused representation is computed as
\vspace{-2pt}
\begin{equation}
F_{\text{AMFB}} = W_{\text{Local}} \cdot F_{\text{Local}} + W_{\text{Global}} \cdot F_{\text{Global}} + W_{\text{Adaptive}} \cdot F_{\text{Adaptive}}.
\label{eq:amfb_final}
\end{equation}

This Tri-stream design offers several advantages. By decoupling the extraction of local details, global context, and adaptive cross-modal cues, the model can dynamically adjust its focus depending on the input. Such flexibility enables the network to prioritize the most informative features for each scenario, thereby enhancing the overall effectiveness of downstream modules.

\subsection{Saliency Decoder}
To decode the enhanced multiscale spatio-temporal features, our hierarchical decoder processes each level independently to fully exploit complementary information. Inspired by recent studies \cite{xiong2023casp, bellitto2021hierarchical}, each decoding block employs 3D convolutions, trilinear upsampling, and a sigmoid activation to generate an intermediate saliency map (see ~\autoref{fig:dtfsal_Network}):
\vspace{-5pt}
\begin{equation}
s_i \in \mathbb{R}^{1 \times 1 \times H \times W} \quad \text{for } i=1,2,3,4. 
\end{equation}
These intermediate outputs capture diverse views of the attention distribution, each reflecting different aspects of the underlying features. To obtain a definitive and coherent saliency prediction, the decoder outputs are concatenated and then fused by a final 3D convolution followed by a sigmoid activation. This fusion step is formalized as:
\vspace{-3pt}
\begin{equation}
	S = \text{Sigmoid}\Bigl(\operatorname{Conv}\Bigl([s_1; s_2; s_3; s_4]\Bigr)\Bigr),
	\label{eq:decoder_final}
\end{equation}

\noindent The learned 3D convolution integrates these diverse perspectives in an optimal way, reducing uncertainty and producing a final saliency map \(S\) that is both precise and coherent.

\subsection{Loss Function}

Our model employs a composite loss function, \(\mathcal{L}_{sal}\), that combines the Kullback-Leibler divergence (KL) and the linear correlation coefficient (CC) to capture both distributional differences and spatial correlations, a strategy inspired by \cite{xiong2023casp, xie2024audio, tsiami2020stavis}.
In this framework, KL quantifies the divergence between the predicted and ground truth saliency distributions, CC measures their linear correlation. This approach thoroughly evaluates prediction errors to effectively optimize the model. The formulation is as follows:
\vspace{-3pt}
\begin{equation}
    \mathcal{L_{\text{KL}}}(s^{g}, s) = \sum_{i=1}^{n} s^{g}_i \log\left(\epsilon + \frac{s^{g}_i}{s_i + \epsilon}\right),
    \label{eq:kl}
\end{equation}
\vspace{-3pt}
\begin{equation}
    \mathcal{L_{\text{CC}}}(s^{g}, s) = \frac{\text{cov}(s^{g}, s)}{\sigma(s^{g}) \cdot \sigma(s)},
    \label{eq:cc}
\end{equation}

\vspace{-3pt}

\begin{equation}
\mathcal{L}_{sal} = \lambda_1 \cdot \mathcal{L_{\text{KL}}}(s^{g}, s)
             + \lambda_2 \cdot \mathcal{L_{\text{CC}}}(s^{g}, s) 
\label{eq:loss}
\end{equation}

\noindent where \( s \) and \( s^{g} \) are the predicted saliency and ground truth (GT) maps, respectively, \( i \) is the pixel index, \( \epsilon \) is a regularization term, and \( \sigma(s, s^{g}) \) is the covariance of \( s \) and \( s^{g} \).
\vspace{-15pt}
\section{Experiments} \label{sec:exp}
\vspace{-5pt}

Experiments are carried out on eight datasets, comprising two purely visual and six audio-visual eye-tracking datasets. The subsequent sections provide details on the implementation and evaluation metrics. Additionally, we present and analyze the experimental results through ablation studies and comparisons with SOTA methods.

\begin{table*}[htbp]
	\begin{center}
		\caption{Comparison with previous methods on six audio-visual saliency datasets. For our model, we indicate the percentage (\%) change in performance relative to the second-best result, or to the best result if ours is not the top performer. The best results are highlighted in \textcolor{red}{red}, the second-best in \textcolor{blue}{blue}, and the third-best in \textcolor{darkgreen}{green}.}
		\label{tab:table-av-sota}
		\vspace*{-5pt}
		\resizebox{\linewidth}{!}{
			\begin{tabular}{c | c | cccc | cccc | cccc}
				\toprule
				\textbf{Method} & \textbf{\#Parameters} & \multicolumn{4}{c|}{\textbf{DIEM}} & \multicolumn{4}{c|}{\textbf{Coutrot1}} & \multicolumn{4}{c}{\textbf{Coutrot2}}\\
				\cline{3-14}
				& & CC $\uparrow$ & NSS $\uparrow$ & AUC-J $\uparrow$  & SIM  $\uparrow$ & CC $\uparrow$ & NSS $\uparrow$ & AUC-J $\uparrow$   & SIM $\uparrow$ & CC $\uparrow$ & NSS $\uparrow$ & AUC-J $\uparrow$  & SIM $\uparrow$ \\
				\hline
				STAViS \cite{tsiami2020stavis} & $20.76M$ & $0.579$ & $2.26$ & $0.883$ & $0.482$ & $0.472$ & $2.11$ & $0.868$ & $0.393$ & $0.734$ & $5.28$ & $0.958$ & $0.511$ \\
				ViNet \cite{jain2021vinet} & $33.97M$ & $0.632$ & $2.53$ & $0.899$ & $0.498$ & $0.560$ & $2.73$ & $0.889$ & $0.425$ & $0.754$ & $5.95$ & $0.951$ & $0.493$ \\
				TSFP-Net \cite{chang2021temporal} & - & $0.651$ & $2.62$ & $0.906$ & $0.527$ & $0.571$ & $2.73$ & $0.895$ & $0.447$ & $0.743$ & $5.31$ & $0.959$ & $0.528$ \\
				CASP-Net \cite{xiong2023casp} & $51.62M$ & $0.655$ & $2.61$ & $0.906$ & $0.543$ & $0.561$ & $2.65$ & $0.889$ & $0.456$ & $0.788$ & $6.34$ & $0.963$ & $0.585$ \\ 
				DiffSal \cite{xiong2024diffsal} & $76.57M$ & $0.660$ & $2.65$ & $0.907$ & $0.543$ & \textcolor{blue}{$0.638$} & \textcolor{darkgreen}{$3.20$} & \textcolor{darkgreen}{$0.901$} & \textcolor{blue}{$0.515$} & $0.835$ & \textcolor{blue}{$6.61$} & $0.964$ & \textcolor{blue}{$0.625$} \\
				MSPI \cite{xie2024audio} & $105.25M$ & $0.671$ & $2.71$ & $0.905$ & $0.545$ & $0.605$ & $2.98$ & $0.899$ & $0.473$ & $0.799$ & $6.40$ & $0.964$ & $0.5961$ \\
				DAVS \cite{zhu2024discrete} & - & $0.580$ & $2.28$ & $0.884$ & $0.483$ & $0.481$ & $2.19$ & $0.869$ & $0.400$ & $0.734$ & $4.984$ & $0.960$ & $0.512$ \\
				RAVF \cite{yu2024relevance} & - & \textcolor{red}{$0.685$} & \textcolor{blue}{$2.78$} & \textcolor{blue}{$0.911$} & \textcolor{blue}{$0.559$} & $0.595$ & $2.87$ & $0.899$ & $0.463$ & \textcolor{darkgreen}{$0.837$} & \textcolor{darkgreen}{$6.47$} & \textcolor{darkgreen}{$0.964$} & $0.617$ \\
                \midrule
				\rowcolor[HTML]{DADCFF} \textbf{DTFSal(V)} & $40.73M$ & \textcolor{darkgreen}{$0.677$} & \textcolor{darkgreen}{$2.76$} & \textcolor{darkgreen}{$0.910$} & \textcolor{darkgreen}{$0.548$} & \textcolor{darkgreen}{$0.623$} & \textcolor{blue}{$3.21$} & \textcolor{blue}{$0.904$} & \textcolor{darkgreen}{$0.505$} & \textcolor{blue}{$0.842$} & $6.42$ & \textcolor{blue}{$0.965$} & \textcolor{darkgreen}{$0.618$} \\
				\rowcolor[HTML]{DADCFF} \textbf{DTFSal} & $49.08M$ & \textcolor{blue}{$0.682$\textsubscript{\scriptsize{-0.44\%}}} & \textcolor{red}{$2.98$\textsubscript{\scriptsize{+6.47\%}}} & \textcolor{red}{$0.921$\textsubscript{\scriptsize{+1.10\%}}} & \textcolor{red}{$0.562$\textsubscript{\scriptsize{+0.54\%}}} & \textcolor{red}{$0.661$\textsubscript{\scriptsize{+3.61\%}}} & \textcolor{red}{$3.31$\textsubscript{\scriptsize{+3.44\%}}} & \textcolor{red}{$0.911$\textsubscript{\scriptsize{+1.11\%}}} & \textcolor{red}{$0.523$\textsubscript{\scriptsize{+1.55\%}}} & \textcolor{red}{$0.845$\textsubscript{\scriptsize{+0.96\%}}} & \textcolor{red}{$6.72$\textsubscript{\scriptsize{+1.66\%}}} & \textcolor{red}{$0.971$\textsubscript{\scriptsize{+0.73\%}}} & \textcolor{red}{$0.649$\textsubscript{\scriptsize{+3.84\%}}} \\
				\hline
				\multirow{2}*{\textbf{Method}}  & \multirow{2}*{\textbf{\#Parameters}} & \multicolumn{4}{c|}{\textbf{AVAD}} & \multicolumn{4}{c|}{\textbf{ETMD}} & \multicolumn{4}{c}{\textbf{SumMe}}\\
				\cline{3-14}
				& & CC $\uparrow$ & NSS $\uparrow$ & AUC-J $\uparrow$  & SIM  $\uparrow$ & CC $\uparrow$ & NSS $\uparrow$ & AUC-J $\uparrow$   & SIM $\uparrow$ & CC $\uparrow$ & NSS $\uparrow$ & AUC-J $\uparrow$  & SIM $\uparrow$ \\
				\hline
				STAViS \cite{tsiami2020stavis} & $20.76M$ & $0.608$ & $3.18$ & $0.919$ & $0.457$ & $0.569$ & $2.94$ & $0.931$ & $0.425$ & $0.422$ & $2.04$ & $0.888$ & $0.337$ \\
				ViNet \cite{jain2021vinet} & $33.97M$ & $0.674$ & $3.77$ & $0.927$ & $0.491$ & $0.571$ & $3.08$ & $0.928$ & $0.406$ & $0.463$ & $2.41$ & $0.897$ & $0.343$ \\
				TSFP-Net \cite{chang2021temporal} & - & $0.704$ & $3.77$ & $0.932$ & $0.521$ & $0.576$ & $3.07$ & $0.932$ & $0.428$ & $0.464$ & $2.30$ & $0.894$ & $0.360$ \\
				CASP-Net \cite{xiong2023casp} & $51.62M$ & $0.691$ & $3.81$ & $0.933$ & $0.528$ & $0.620$ & $3.34$ & \textcolor{darkgreen}{$0.940$} & $0.478$ & $0.499$ & $2.60$ & $0.907$ & $0.387$ \\ 
				DiffSal \cite{xiong2024diffsal} & $76.57M$ & \textcolor{darkgreen}{$0.738$} & \textcolor{blue}{$4.22$} & $0.935$ & \textcolor{blue}{$0.571$} & \textcolor{darkgreen}{$0.652$} & \textcolor{blue}{$3.66$} & \textcolor{blue}{$0.943$} & \textcolor{blue}{$0.506$} & \textcolor{blue}{$0.572$} & \textcolor{blue}{$3.14$} & \textcolor{blue}{$0.921$} & \textcolor{blue}{$0.447$} \\
				MSPI  \cite{xie2024audio} & $105.25M$ & $0.702$ & $3.92$ & \textcolor{darkgreen}{$0.936$} & $0.533$ & $0.623$ & $3.35$ & $0.940$ & $0.471$ & $0.502$ & $2.61$ & $0.909$ & $0.380$ \\
				DAVS  \cite{zhu2024discrete} & - & $0.610$ & $3.19$ & $0.919$ & $0.458$ & $0.600$ & $2.96$ & $0.932$ & $0.425$ & $0.422$ & $2.28$ & $0.888$ & $0.338$ \\
				RAVF \cite{yu2024relevance} & - & $0.704$ & $3.84$ & $0.933$ & $0.549$ & $0.611$ & $3.36$ & $0.934$ & $0.474$ & $0.504$ & $2.72$ & $0.903$ & $0.381$ \\
				\midrule
				\rowcolor[HTML]{DADCFF} \textbf{DTFSal(V)} & $40.73M$ & \textcolor{blue}{$0.742$} & \textcolor{darkgreen}{$4.21$} & \textcolor{blue}{$0.938$} & \textcolor{darkgreen}{$0.559$} & \textcolor{blue}{$0.653$} & \textcolor{darkgreen}{$3.57$} & $0.939$ & \textcolor{darkgreen}{$0.499$} & \textcolor{darkgreen}{$0.561$} & \textcolor{darkgreen}{$2.91$} & \textcolor{darkgreen}{$0.911$} & \textcolor{darkgreen}{$0.441$} \\
				\rowcolor[HTML]{DADCFF} \textbf{DTFSal} & $49.08M$ & \textbf{\textcolor{red}{$0.746$\textsubscript{\scriptsize{+1.09\%}}} }& \textcolor{red}{$4.36$\textsubscript{\scriptsize{+3.32\%}}} & \textcolor{red}{$0.939$\textsubscript{\scriptsize{+0.32\%}}} & \textcolor{red}{$0.595$\textsubscript{\scriptsize{+4.64\%}}} & \textcolor{red}{$0.667$\textsubscript{\scriptsize{+2.30\%}}} & \textcolor{red}{$3.82$\textsubscript{\scriptsize{+4.37\%}}} & \textcolor{red}{$0.957$\textsubscript{\scriptsize{+1.48\%}}} & \textcolor{red}{$0.531$\textsubscript{\scriptsize{+4.94\%}}} & \textcolor{red}{$0.573$\textsubscript{\scriptsize{+0.175\%}}} & \textcolor{red}{$3.20$\textsubscript{\scriptsize{+1.91\%}}} & \textcolor{red}{$0.927$\textsubscript{\scriptsize{+0.65\%}}} & \textcolor{red}{$0.449$\textsubscript{\scriptsize{+0.45\%}}} \\
				\bottomrule
		\end{tabular}
		}
	\end{center}
	\vspace{-10pt}
\end{table*}

\subsection{Setup}

\noindent \textbf{Datasets:} We utilize DHF1K \cite{Wang2018revisiting} and UCF-Sports \cite{peng2018two} as visual datasets, and DIEM \cite{mital2011clustering}, ETMD \cite{koutras2015perceptually}, SumMe \cite{gygli2014creating}, AVAD \cite{min2016fixation}, Coutrot1 \cite{coutrot2014saliency}, and Coutrot2 \cite{coutrot2016multimodal} as audio-visual datasets. For further details, please refer to the \textcolor{red}{Supplementary Material}.

\vspace{3pt}

\noindent \textbf{Implementation Details:}
Our model is implemented using the PyTorch framework and is trained on an NVIDIA GeForce RTX 4090 GPU for 10 epochs with early stopping after 3 epochs. We adopt a pre-trained MViTv2~\cite{li2022mvitv2} model on Kinetics~\cite{carreira2017quo} and a pre-trained VGGish~\cite{hershey2017cnn} on AudioSet~\cite{gemmeke2017audio} as the visual and audio encoders, respectively; the visual branch of DTFSal is pre-trained on the DHF1k dataset following \cite{xiong2024diffsal, xiong2023casp}, and the entire model is then fine-tuned on the target audio-visual datasets using the pre-trained weights. The input samples to the network consist of 32-frame video clips of size 224 × 384 × 3, accompanied by their corresponding audio, which is transformed into 9 slices of 112 × 192 log-Mel spectrograms. For training, the Adam optimizer is utilized with an initial learning rate set to $1 \times 10^{-4}$, and we employ a learning rate scheduler that decreases the factor by 0.1 every three epochs. The batch size is set to 1, and LN is applied to avoid gradient explosion and offset the effects of a small batch size \cite{xie2024audio, jin2025hierarchical}. In addition, the optimal values for the loss function's weighting coefficients are set as $\lambda_1 = 1$ and $\lambda_2 = -1$.

\vspace{3pt}

\noindent \textbf{Evaluation Metrics:}
We employed four standard evaluation metrics: CC, NSS (Normalized Scan-path Saliency), AUC-J (Area Under Curve - Judd), and SIM (Similarity) to compare our work with existing studies \cite{xiong2023casp, tan2024transformer, tsiami2020stavis}. The SIM assesses the overlap between the predicted and actual saliency distributions. The AUC-J is a location-based metric that evaluates the performance of the predicted saliency map as a binary classifier relative to the true saliency map. Meanwhile, NSS measures the average normalized saliency at the specific locations of human eye fixations. For more detailed information on saliency metrics, please refer to \cite{bylinskii2018different}.

\vspace{-3pt}

\begin{figure*}[t]
    \centering
    \includegraphics[width=\textwidth]{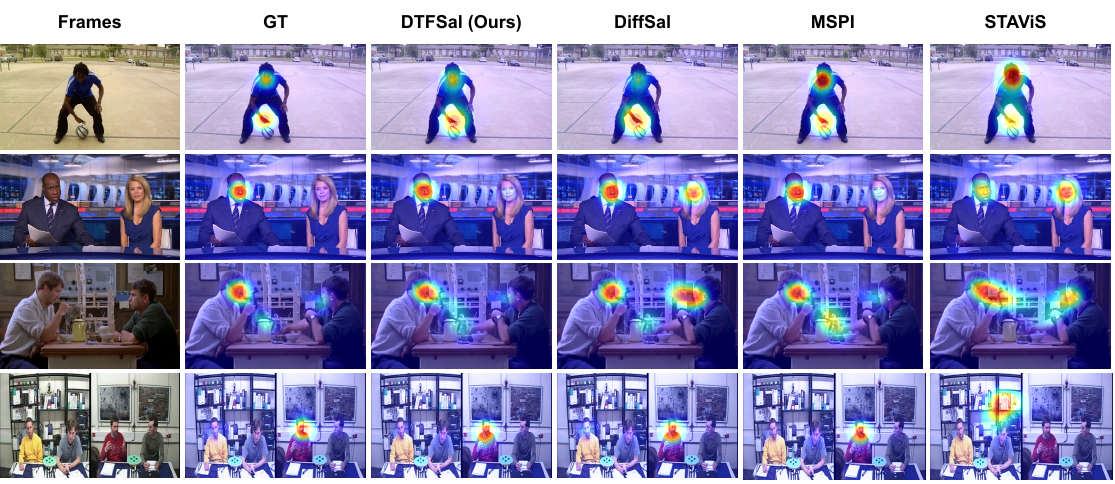}
    \caption{Comparative visualizations of our DTFSal model compared with previous SOTA audio-visual saliency prediction methods.}
    \label{fig:qualitative_result_fig_AV}
\end{figure*}

\subsection{Method Comparison}

\vspace{-3pt}

\noindent \textbf{Comparisons with SOTA Methods:} As shown in ~\autoref{tab:table-av-sota}, the experimental results of our DTFSal are compared with recent SOTA methods on six audio-visual saliency datasets. The table highlights the superiority of DTFSal, as it outperforms all comparable methods on all datasets—achieving the best performance in 23 out of 24 metrics and ranking second in only one—thus establishing a new SOTA on these benchmarks. Notably, DTFSal significantly surpasses previous top-performing methods, such as CASP-Net~\cite{xiong2023casp} and DiffSal~\cite{xiong2024diffsal}, while utilizing fewer parameters and incurring lower computational costs. Although DiffSal can achieve an average relative performance improvement of up to 6.3\% compared to the second-best method~\cite{xiong2024diffsal}, our DTFSal, with its more efficient design, attains an average relative performance improvement of up to \textbf{2.1\%} over the second-place results, including those of DiffSal and other SOTA models. Additionally, our visual-only variant, denoted as DTFSal(V), achieves competitive results on audio-visual datasets, further demonstrating the impact of our AMFB module.

\vspace{3pt}

\noindent \textbf{Comparison with Video Saliency prediction methods:} ~\autoref{tab:comparison} in \textcolor{red}{Supplementary} compares our visual-only variant, DTFSal(V), with recent SOTA video saliency prediction methods on the DHF1K and UCF Sports datasets. Our results show that across these two datasets, DTFSal(V) achieves the best performance in 7 metrics and ranks second in the remaining one. Moreover, the relative performance improvement over the second-best method is 2.06\% on these datasets, indicating that even our visual-only variant can reach SOTA performance when compared to methods such as ViNet-E~\cite{girmaji2025minimalistic} and HSFI-Net~\cite{jin2025hierarchical}.

\begin{table}[htbp]
\centering
\caption{Comparison of FLOPs, input size, and model size between ours and previous visual and audio-visual saliency methods.}
\label{tab:model_comparison_flops}
\resizebox{0.85\columnwidth}{!}{%
\begin{tabular}{lccc}
\toprule
\textbf{Model} & \textbf{Input Size} & \textbf{FLOPs (G)} & \textbf{Model Size (MB)} \\
\midrule
\multicolumn{4}{c}{\textbf{Visual Models}} \\
\midrule
DeepVS \cite{jiang2018deepvs}    & 448$\times$448 & 819.5 & 344 \\
STRA-Net \cite{lai2019video}  & 256$\times$224 & 55    & 324 \\
TASED-Net \cite{min2019tased} & 256$\times$224 & 91.5  & 82  \\
UNISAL  \cite{droste2020unified}  & 288$\times$224 & 14.4  & 146 \\
HD2S   \cite{bellitto2021hierarchical}    & 192$\times$128 & 22    & 155 \\
ViNet   \cite{jain2021vinet}    & 192$\times$128 & 115   & 152 \\
STSA-Net   \cite{wang2021spatio} & 384$\times$224 & 493   & 245 \\
TMFI-Net  \cite{zhou2023transformer} & 384$\times$224 & 358   & 234 \\
THTD-Net \cite{moradi2024transformer} & 384$\times$224 & -  & 220 \\
TM2SP-Net \cite{li2025tm2sp} & 384$\times$224 & 1330  & 700 \\
ViNet-E \cite{girmaji2025minimalistic} & 384$\times$224 & -  & 183.84 \\
HSFI-Net \cite{jin2025hierarchical} & 384$\times$224 & -  & 253 \\
DTFSal(V) & 384$\times$224 & 259.57  & 155.37 \\
\midrule
\multicolumn{4}{c}{\textbf{Audio-Visual Models}} \\
\midrule
ACLNet   \cite{wang2019revisiting}  & 224$\times$224 & 164   & 250 \\
TSFP-Net \cite{chang2021temporal} & 352$\times$192 & 316   & 192 \\
STAVis \cite{tsiami2020stavis} & 384$\times$224 & 15.31  & 79.19 \\
Casp-Net \cite{xiong2023casp} & 384$\times$224 & 283.35  & 196.91 \\
DiffSal \cite{xiong2024diffsal} & 384$\times$224 & 834  & 351 \\
DTFSal & 384$\times$224 & 297.91  & 187.26 \\
\bottomrule
\end{tabular}}
\end{table}


\noindent \textbf{Qualitative Results:}~\autoref{fig:qualitative_result_fig_AV} presents visual comparisons between our DTFSal model and recent audio-visual saliency prediction methods. The saliency maps generated by DTFSal align more closely with the GT, clearly highlighting key regions and accurately outlining their boundaries (For more visualization results see \textcolor{red}{Supplementary}).

\vspace{5pt}

\noindent \textbf{Efficiency Analysis:}~\autoref{tab:model_comparison_flops} presents a comparison of computational complexity and model size for both visual and audio-visual saliency models. Our proposed DTFSal achieves a balanced trade-off between efficiency and effectiveness, requiring 297.91 GFLOPs with a model size of 187.26 MB, which is notably more compact than DiffSal while achieving better performance. Similarly, its visual-only counterpart, DTFSal(V), maintains a moderate GFLOPs, outperforming several transformer-based visual models with excessive computational demands, such as TM2SP-Net. Despite its efficiency, DTFSal delivers strong performance across multiple saliency benchmarks, as shown in~\autoref{tab:table-av-sota}, where it obtains SOTA results. With a parameter count of 49.08M, DTFSal integrates multimodal information, demonstrating its capability to use audio-visual cues for enhanced saliency prediction while maintaining an optimal balance between accuracy and efficiency.


\noindent \textbf{Role of Audio Information:} Our analysis of audio cues in audio-visual saliency prediction involves experiments on several benchmark datasets. We evaluated the model's performance by disabling the audio feature extraction branch while keeping all other parameters unchanged. As shown in \autoref{tab:table-av-sota} and ~\autoref{tab:table-v-sota} in the \textcolor{red}{Supplementary Material}, incorporating audio cues consistently enhances saliency prediction across all six datasets. This improvement is driven by audio features' ability to accurately localize sound-emitting objects and capture key cross-modal relationships between auditory and visual stimuli \cite{zhu2024does, zhu2024discrete}. Notably, our DTFSal(V) model achieves competitive or superior performance compared to SOTA methods, highlighting the importance of integrating audio information for refined attention mechanisms and enhanced prediction accuracy.

\begin{table}[!tbp]
	\caption{Performance comparison of block types (LTEB and DLTFB) across various stage combinations on DHF1K dataset.}
	\resizebox{\linewidth}{!}{%
	\begin{tabular}{llccccccc}
		\toprule
		\multirow{2}{*}{Block Type} & \multirow{2}{*}{Metric} & \multicolumn{7}{c}{$i$-th stage of Video Encoder, $i \in \{1, 2, 3, 4\}$} \\
		\cmidrule(lr){3-9}
		& & No Block & 1 & 2 & 3 & 4 & 2,3,4 & 1,2,3,4 \\
		\midrule\midrule
		\multirow{4}{*}{LTEB} 
		  & CC    & $0.539$      & $0.547$ & $0.551$ & $0.556$ & $0.556$ & $0.589$ & \cellcolor[HTML]{DADCFF}\textbf{0.560} \\
		  & NSS   & $2.972$      & $3.080$ & $3.081$ & $3.133$ & $3.117$ & $3.132$ & \cellcolor[HTML]{DADCFF}\textbf{3.190} \\
		  & AUC-J & $0.911$      & $0.913$ & $0.916$ & $0.919$ & $0.918$ & $0.920$ & \cellcolor[HTML]{DADCFF}\textbf{0.922} \\
		  & SIM   & $0.432$      & $0.438$ & $0.435$ & $0.436$ & $0.434$ & $0.434$ & \cellcolor[HTML]{DADCFF}\textbf{0.436} \\
		\midrule
		\multirow{4}{*}{DLTFB} 
		  & CC    & $-$      & $0.557$ & $0.549$ & $0.560$ & \cellcolor[HTML]{DADCFF}\textbf{0.561} & $-$     & $0.560$ \\
		  & NSS   & $-$      & $3.148$ & $3.108$ & $3.190$ & \cellcolor[HTML]{DADCFF}\textbf{3.205} & $-$     & $3.149$ \\
		  & AUC-J & $-$      & $0.920$ & $0.918$ & $0.922$ & \cellcolor[HTML]{DADCFF}\textbf{0.923} & $-$     & $0.922$ \\
		  & SIM   & $-$      & $0.439$ & $0.433$ & $0.436$ & \cellcolor[HTML]{DADCFF}\textbf{0.446} & $-$     & $0.432$ \\
		\bottomrule
	\end{tabular}%
	}
	\label{table-block-type-ablation}
\end{table}

\subsection{Ablation Studies}
\vspace{-3pt}
To validate our design choices, comprehensive ablation studies are conducted. We follow the evaluation protocols in~\cite{xiong2024diffsal, xiong2023casp, yu2024relevance} in the AVAD and ETMD datasets for audio-visual evaluation, while the DHF1K dataset is additionally included to evaluate our visual baseline. Due to unavailable annotations on the DHF1K test set, the DFTSal model is evaluated solely on its validation set, following the approach in \cite{xiong2024diffsal, hu2023tinyhd}.


\begin{table}[bp]
\centering
\caption{Comparison of Different Multi-Modal Feature Fusion Methods.}
\label{tab:fusion-methods}
\resizebox{\linewidth}{!}{%
\begin{tabular}{lcccccccc}
\toprule
\multirow{2}{*}{Fusion Methods} & \multicolumn{4}{c}{ETMD} & \multicolumn{4}{c}{AVAD} \\
\cmidrule(lr){2-5}\cmidrule(lr){6-9}
 & SIM$\uparrow$ & CC$\uparrow$ & NSS$\uparrow$ & AUC-J$\uparrow$ & SIM$\uparrow$ & CC$\uparrow$ & NSS$\uparrow$ & AUC-J$\uparrow$ \\
\midrule
Concatenation              & 0.381 & 0.625 & 2.74 & 0.910 & 0.531 & 0.690 & 3.61  & 0.920 \\
Cross-attention                   & 0.485 & 0.635 & 3.16 & 0.907 & 0.539 & 0.696 & 3.563  & 0.926 \\
\rowcolor[HTML]{DADCFF} AMFB (Ours)                & $\textbf{0.531}$ & $\textbf{0.667}$ & $\textbf{3.82}$ & $\textbf{0.957}$ & $\textbf{0.595}$ & $\textbf{0.746}$ & $\textbf{4.36}$  & $\textbf{0.939}$ \\
\bottomrule
\end{tabular}%
}
\end{table}

\vspace{3pt}

\noindent \textbf{Feature Enhancement Block Performance Across Stages:}~\autoref{table-block-type-ablation} shows the performance of our modules on DHF1K dataset when applied at different stages of the encoder. To assess the effectiveness of the LTEB module, we evaluate three configurations: a baseline model without LTEB, integrating it at a single stage, and deploying it across multiple stages. The results indicate that incorporating LTEB improves all metrics, with the best performance achieved when LTEB is applied at all stages. Similarly, our evaluation of the DLTFB module reveals that employing dynamic spatial shifting exclusively at stage 4 yields better results compared to other placements, highlighting the advantage of optimizing the spatial token arrangement in the final stage. These ablation experiments validate our design choices and demonstrate that both LTEB and DLTFB enhance spatio–temporal feature refinement.


\vspace{3pt}

\noindent \textbf{Impact of Various Multi-Modal Feature Fusion Methods:}
We evaluate our proposed Tri-stream, AMFB, against two alternative fusion methods: simple concatenation of audio and visual features, and a cross-attention mechanism that uses visual features as queries and values and audio features as keys, similar to the approach in~\cite{xiong2024diffsal}. As shown in ~\autoref{tab:fusion-methods}, AMFB outperforms these alternatives on both the ETMD and AVAD datasets, achieving higher scores. This improvement is attributed to its Tri-stream design, which effectively captures local details, global context, and adaptive cross-modal information, leading to a more precise fusion of complementary cues.

\vspace{3pt}

\noindent \textbf{Effectiveness of Multi-Decoder in Saliency Prediction:} To validate our multi-decoder design, we compare it with two alternative decoder architectures: one that concatenates all features and uses a single decoder based on 3D convolutions and upsampling, and a UNet-style decoder that progressively upsamples using alternating 3D convolutions and trilinear layers with skip connections. As shown in~\autoref{table-decoder}, our multi-decoder approach achieves superior performance on DHF1K \cite{Wang2018revisiting}, confirming its effectiveness in refining spatio–temporal features for accurate saliency prediction.

\begin{table}[tbp]
	\caption{Comparison between Multi Decoders and other decoders on DHF1K. \#Params represent the number of parameters per decoder.}
	\resizebox{\linewidth}{!}{%
		\begin{tabular}{l|ccccc}
        \hline
			\multicolumn{1}{c|}{Method}  & \multicolumn{1}{c}{\#Params} & CC$\uparrow$ & NSS$\uparrow$  & SIM$\uparrow$ & AUC-J$\uparrow$ \\
			\hline
			One Decoder         & \textbf{0.43} & 0.558  & 3.148  & 0.425  & 0.920 \\
			U-Net      & 2.84          & 0.555  & 3.154  & 0.414  & 0.922 \\
			\rowcolor[HTML]{DADCFF} \textbf{Multi-Decoder} & 1.74       & \textbf{0.561}  & \textbf{3.205}  & \textbf{0.436}  & \textbf{0.923} \\
            \hline
		\end{tabular}%
	}
	\label{table-decoder}
\end{table}

\vspace{3pt}

\noindent \textbf{Additional Ablation Studies:}
For further analysis of our approach, please refer to \textcolor{red}{Supplementary Material}.

\section{Conclusion} \label{sec:Conclusion}
In this paper, we presented DFTSAL, a novel framework for dynamic video saliency prediction that balances performance and efficiency through innovative modules such as LTEB, DLTFB, and a Tri-stream AMFB for effective audio-visual integration. Our approach robustly captures multi-scale spatio–temporal features and effectively fuses audio cues, achieving SOTA performance on both visual-only and audio-visual benchmarks with competitive computational efficiency. Experimental results and extensive ablation studies demonstrate the effectiveness of our method, paving the way for scalable applications in real-world scenarios.

{
    \small
    \bibliographystyle{ieeenat_fullname}
    \bibliography{main}

\begin{thebibliography}{69}
\providecommand{\natexlab}[1]{#1}
\providecommand{\url}[1]{\texttt{#1}}
\expandafter\ifx\csname urlstyle\endcsname\relax
  \providecommand{\doi}[1]{doi: #1}\else
  \providecommand{\doi}{doi: \begingroup \urlstyle{rm}\Url}\fi

\bibitem[Aydemir et~al.(2023)Aydemir, Hoffstetter, Zhang, Salzmann, and S\"usstrunk]{aydemir2023tempsal}
Bahar Aydemir, Ludo Hoffstetter, Tong Zhang, Mathieu Salzmann, and Sabine S\"usstrunk.
\newblock Tempsal - uncovering temporal information for deep saliency prediction.
\newblock In \emph{Proceedings of the IEEE/CVF Conference on Computer Vision and Pattern Recognition (CVPR)}, pages 6461--6470, 2023.

\bibitem[Bartolo and Seychell(2024)]{bartolo2024correlation}
Matthias Bartolo and Dylan Seychell.
\newblock Correlation of object detection performance with visual saliency and depth estimation.
\newblock \emph{arXiv preprint arXiv:2411.02844}, 2024.

\bibitem[Bellitto et~al.(2021)Bellitto, Proietto~Salanitri, Palazzo, Rundo, Giordano, and Spampinato]{bellitto2021hierarchical}
Giovanni Bellitto, Federica Proietto~Salanitri, Simone Palazzo, Francesco Rundo, Daniela Giordano, and Concetto Spampinato.
\newblock Hierarchical domain-adapted feature learning for video saliency prediction.
\newblock \emph{International Journal of Computer Vision}, 129:\penalty0 3216--3232, 2021.

\bibitem[Bruckert et~al.(2021)Bruckert, Tavakoli, Liu, Christie, and Le~Meur]{bruckert2021deep}
Alexandre Bruckert, Hamed~R Tavakoli, Zhi Liu, Marc Christie, and Olivier Le~Meur.
\newblock Deep saliency models: The quest for the loss function.
\newblock \emph{Neurocomputing}, 453:\penalty0 693--704, 2021.

\bibitem[Bylinskii et~al.(2018)Bylinskii, Judd, Oliva, Torralba, and Durand]{bylinskii2018different}
Zoya Bylinskii, Tilke Judd, Aude Oliva, Antonio Torralba, and Fr{\'e}do Durand.
\newblock What do different evaluation metrics tell us about saliency models?
\newblock \emph{IEEE transactions on pattern analysis and machine intelligence}, 41\penalty0 (3):\penalty0 740--757, 2018.

\bibitem[Carreira and Zisserman(2017)]{carreira2017quo}
Joao Carreira and Andrew Zisserman.
\newblock Quo vadis, action recognition? a new model and the kinetics dataset.
\newblock In \emph{proceedings of the IEEE Conference on Computer Vision and Pattern Recognition}, pages 6299--6308, 2017.

\bibitem[Chang and Zhu(2021)]{chang2021temporal}
Qinyao Chang and Shiping Zhu.
\newblock Temporal-spatial feature pyramid for video saliency detection.
\newblock \emph{arXiv preprint arXiv:2105.04213}, 2021.

\bibitem[Chang and Zhu(2023)]{Chang2023}
Qinyao Chang and Shiping Zhu.
\newblock Human vision attention mechanism-inspired temporal-spatial feature pyramid for video saliency detection.
\newblock \emph{Cognitive Computation}, 15\penalty0 (3):\penalty0 856--868, 2023.

\bibitem[Coutrot and Guyader(2014)]{coutrot2014saliency}
Antoine Coutrot and Nathalie Guyader.
\newblock How saliency, faces, and sound influence gaze in dynamic social scenes.
\newblock \emph{Journal of Vision}, 14\penalty0 (8):\penalty0 5--5, 2014.

\bibitem[Coutrot and Guyader(2016)]{coutrot2016multimodal}
Antoine Coutrot and Nathalie Guyader.
\newblock Multimodal saliency models for videos.
\newblock \emph{From Human Attention to Computational Attention: A Multidisciplinary Approach}, pages 291--304, 2016.

\bibitem[Dai et~al.(2017)Dai, Qi, Xiong, Li, Zhang, Hu, and Wei]{dai2017deformable}
Jifeng Dai, Haozhi Qi, Yuwen Xiong, Yi Li, Guodong Zhang, Han Hu, and Yichen Wei.
\newblock Deformable convolutional networks.
\newblock In \emph{Proceedings of the IEEE international conference on computer vision}, pages 764--773, 2017.

\bibitem[Das et~al.(2024)Das, Wu, Skrjanec, and Feit]{das2024shifting}
Anwesha Das, Zekun Wu, Iza Skrjanec, and Anna~Maria Feit.
\newblock Shifting focus with hceye: Exploring the dynamics of visual highlighting and cognitive load on user attention and saliency prediction.
\newblock \emph{Proceedings of the ACM on Human-Computer Interaction}, 8\penalty0 (ETRA):\penalty0 1--18, 2024.

\bibitem[Droste et~al.(2020)Droste, Jiao, and Noble]{droste2020unified}
Richard Droste, Jianbo Jiao, and J~Alison Noble.
\newblock Unified image and video saliency modeling.
\newblock In \emph{Computer Vision--ECCV 2020: 16th European Conference, Glasgow, UK, August 23--28, 2020, Proceedings, Part V 16}, pages 419--435. Springer, 2020.

\bibitem[Fan et~al.(2021)Fan, Xiong, Mangalam, Li, Yan, Malik, and Feichtenhofer]{fan2021multiscale}
Haoqi Fan, Bo Xiong, Karttikeya Mangalam, Yanghao Li, Zhicheng Yan, Jitendra Malik, and Christoph Feichtenhofer.
\newblock Multiscale vision transformers.
\newblock In \emph{Proceedings of the IEEE/CVF international conference on computer vision}, pages 6824--6835, 2021.

\bibitem[Gemmeke et~al.(2017)Gemmeke, Ellis, Freedman, Jansen, Lawrence, Moore, Plakal, and Ritter]{gemmeke2017audio}
Jort~F Gemmeke, Daniel~PW Ellis, Dylan Freedman, Aren Jansen, Wade Lawrence, R~Channing Moore, Manoj Plakal, and Marvin Ritter.
\newblock Audio set: An ontology and human-labeled dataset for audio events.
\newblock In \emph{2017 IEEE International Conference on Acoustics, Speech and Signal Processing (ICASSP)}, pages 776--780. IEEE, 2017.

\bibitem[Girmaji et~al.(2025)Girmaji, Jain, Beri, Bansal, and Gandhi]{girmaji2025minimalistic}
Rohit Girmaji, Siddharth Jain, Bhav Beri, Sarthak Bansal, and Vineet Gandhi.
\newblock Minimalistic video saliency prediction via efficient decoder \& spatio temporal action cues.
\newblock \emph{arXiv preprint arXiv:2502.00397}, 2025.

\bibitem[G{\"u}ng{\"o}rd{\"u} and Tekalp(2024)]{gungordu2024saliency}
O{\u{g}}uzhan G{\"u}ng{\"o}rd{\"u} and A~Murat Tekalp.
\newblock Saliency-aware end-to-end learned variable-bitrate 360-degree image compression.
\newblock In \emph{2024 IEEE International Conference on Image Processing (ICIP)}, pages 1795--1801. IEEE, 2024.

\bibitem[Gygli et~al.(2014)Gygli, Grabner, Riemenschneider, and Van~Gool]{gygli2014creating}
Michael Gygli, Helmut Grabner, Hayko Riemenschneider, and Luc Van~Gool.
\newblock Creating summaries from user videos.
\newblock In \emph{Computer Vision--ECCV 2014: 13th European Conference, Zurich, Switzerland, September 6-12, 2014, Proceedings, Part VII 13}, pages 505--520. Springer, 2014.

\bibitem[Hershey et~al.(2017)Hershey, Chaudhuri, Ellis, Gemmeke, Jansen, Moore, Plakal, Platt, Saurous, Seybold, et~al.]{hershey2017cnn}
Shawn Hershey, Sourish Chaudhuri, Daniel~PW Ellis, Jort~F Gemmeke, Aren Jansen, R~Channing Moore, Manoj Plakal, Devin Platt, Rif~A Saurous, Bryan Seybold, et~al.
\newblock Cnn architectures for large-scale audio classification.
\newblock In \emph{2017 ieee international conference on acoustics, speech and signal processing (icassp)}, pages 131--135. IEEE, 2017.

\bibitem[Hosseini et~al.(2024{\natexlab{a}})Hosseini, Hooshanfar, Omrani, Toosi, Toosi, Ebrahimian, and Akhaee]{hosseini2024brand}
Alireza Hosseini, Kiana Hooshanfar, Pouria Omrani, Reza Toosi, Ramin Toosi, Zahra Ebrahimian, and Mohammad~Ali Akhaee.
\newblock Brand visibility in packaging: A deep learning approach for logo detection, saliency-map prediction, and logo placement analysis.
\newblock \emph{arXiv preprint arXiv:2403.02336}, 2024{\natexlab{a}}.

\bibitem[Hosseini et~al.(2024{\natexlab{b}})Hosseini, Kazerouni, Akhavan, Brudno, and Taati]{hosseini2024sum}
Alireza Hosseini, Amirhossein Kazerouni, Saeed Akhavan, Michael Brudno, and Babak Taati.
\newblock Sum: Saliency unification through mamba for visual attention modeling.
\newblock \emph{arXiv preprint arXiv:2406.17815}, 2024{\natexlab{b}}.

\bibitem[Hu et~al.(2023)Hu, Palazzo, Salanitri, Bellitto, Moradi, Spampinato, and McGuinness]{hu2023tinyhd}
Feiyan Hu, Simone Palazzo, Federica~Proietto Salanitri, Giovanni Bellitto, Morteza Moradi, Concetto Spampinato, and Kevin McGuinness.
\newblock Tinyhd: Efficient video saliency prediction with heterogeneous decoders using hierarchical maps distillation.
\newblock In \emph{Proceedings of the IEEE/CVF Winter Conference on Applications of Computer Vision}, pages 2051--2060, 2023.

\bibitem[Jain et~al.(2021)Jain, Yarlagadda, Jyoti, Karthik, Subramanian, and Gandhi]{jain2021vinet}
Samyak Jain, Pradeep Yarlagadda, Shreyank Jyoti, Shyamgopal Karthik, Ramanathan Subramanian, and Vineet Gandhi.
\newblock Vinet: Pushing the limits of visual modality for audio-visual saliency prediction.
\newblock In \emph{2021 IEEE/RSJ International Conference on Intelligent Robots and Systems (IROS)}, pages 3520--3527. IEEE, 2021.

\bibitem[Jarimijafarbigloo et~al.(2024)Jarimijafarbigloo, Azad, Kazerouni, and Merhof]{jarimijafarbigloo2024reducing}
Sanaz Jarimijafarbigloo, Reza Azad, Amirhossein Kazerouni, and Dorit Merhof.
\newblock Reducing uncertainty in 3d medical image segmentation under limited annotations through contrastive learning.
\newblock \emph{Medical Imaging with Deep Learning}, pages 694--707, 2024.

\bibitem[Jiang et~al.(2018)Jiang, Xu, Liu, Qiao, and Wang]{jiang2018deepvs}
Lai Jiang, Mai Xu, Tie Liu, Minglang Qiao, and Zulin Wang.
\newblock Deepvs: A deep learning based video saliency prediction approach.
\newblock In \emph{Proceedings of the European Conference on Computer Vision}, pages 602--617, 2018.

\bibitem[Jiang et~al.(2023)Jiang, Leiva, Rezazadegan~Tavakoli, RB~Houssel, Kylm{\"a}l{\"a}, and Oulasvirta]{jiang2023ueyes}
Yue Jiang, Luis~A Leiva, Hamed Rezazadegan~Tavakoli, Paul RB~Houssel, Julia Kylm{\"a}l{\"a}, and Antti Oulasvirta.
\newblock Ueyes: Understanding visual saliency across user interface types.
\newblock In \emph{Proceedings of the 2023 CHI Conference on Human Factors in Computing Systems}, pages 1--21, 2023.

\bibitem[Jin et~al.(2025)Jin, Zhou, Zhang, Fang, Shi, and Xu]{jin2025hierarchical}
Yingjie Jin, Xiaofei Zhou, Zhenjie Zhang, Hao Fang, Ran Shi, and Xiaobin Xu.
\newblock Hierarchical spatiotemporal feature interaction network for video saliency prediction.
\newblock \emph{Image and Vision Computing}, page 105413, 2025.

\bibitem[Kocak et~al.(2021)Kocak, Erdem, and Erdem]{kocak2021gated}
Aysun Kocak, Erkut Erdem, and Aykut Erdem.
\newblock A gated fusion network for dynamic saliency prediction.
\newblock \emph{IEEE Transactions on Cognitive and Developmental Systems}, 14\penalty0 (3):\penalty0 995--1008, 2021.

\bibitem[Koch et~al.(2006)Koch, McLean, Segev, Freed, Berry, Balasubramanian, and Sterling]{koch2006much}
Kristin Koch, Judith McLean, Ronen Segev, Michael~A Freed, Michael~J Berry, Vijay Balasubramanian, and Peter Sterling.
\newblock How much the eye tells the brain.
\newblock \emph{Current biology}, 16\penalty0 (14):\penalty0 1428--1434, 2006.

\bibitem[Koutras and Maragos(2015)]{koutras2015perceptually}
Petros Koutras and Petros Maragos.
\newblock A perceptually based spatio-temporal computational framework for visual saliency estimation.
\newblock \emph{Signal Processing: Image Communication}, 38:\penalty0 15--31, 2015.

\bibitem[Lai et~al.(2019)Lai, Wang, Sun, and Shen]{lai2019video}
Qiuxia Lai, Wenguan Wang, Hanqiu Sun, and Jianbing Shen.
\newblock Video saliency prediction using spatiotemporal residual attentive networks.
\newblock \emph{IEEE Transactions on Image Processing}, 29:\penalty0 1113--1126, 2019.

\bibitem[Lazaridis et~al.(2024)Lazaridis, Georgiadis, Kalaganis, Kordopatis-Zilos, Papadopoulos, Nikolopoulos, and Kompatsiaris]{lazaridis2024visual}
Nikos Lazaridis, Kostas Georgiadis, Fotis Kalaganis, Giorgos Kordopatis-Zilos, Symeon Papadopoulos, Spiros Nikolopoulos, and Ioannis Kompatsiaris.
\newblock The visual saliency transformer goes temporal: Tempvst for video saliency prediction.
\newblock \emph{IEEE Access}, 2024.

\bibitem[Li and Liu(2025)]{li2025tm2sp}
Chenming Li and Shiguang Liu.
\newblock Tm2sp: A transformer-based multi-level spatiotemporal feature pyramid network for video saliency prediction.
\newblock \emph{IEEE Transactions on Circuits and Systems for Video Technology}, 2025.

\bibitem[Li et~al.(2022)Li, Wu, Fan, Mangalam, Xiong, Malik, and Feichtenhofer]{li2022mvitv2}
Yanghao Li, Chao-Yuan Wu, Haoqi Fan, Karttikeya Mangalam, Bo Xiong, Jitendra Malik, and Christoph Feichtenhofer.
\newblock Mvitv2: Improved multiscale vision transformers for classification and detection.
\newblock In \emph{Proceedings of the IEEE/CVF Conference on Computer Vision and Pattern Recognition}, pages 4804--4814, 2022.

\bibitem[Liu et~al.(2020)Liu, Qiao, Xu, Li, Hu, and Borji]{liu2020learning}
Yufan Liu, Minglang Qiao, Mai Xu, Bing Li, Weiming Hu, and Ali Borji.
\newblock Learning to predict salient faces: A novel visual-audio saliency model.
\newblock In \emph{Computer Vision--ECCV 2020: 16th European Conference, Glasgow, UK, August 23--28, 2020, Proceedings, Part XX 16}, pages 413--429. Springer, 2020.

\bibitem[Liu et~al.(2021)Liu, Lin, Cao, Hu, Wei, Zhang, Lin, and Guo]{liu2021swin}
Ze Liu, Yutong Lin, Yue Cao, Han Hu, Yixuan Wei, Zheng Zhang, Stephen Lin, and Baining Guo.
\newblock Swin transformer: Hierarchical vision transformer using shifted windows.
\newblock In \emph{Proceedings of the IEEE/CVF international conference on computer vision}, pages 10012--10022, 2021.

\bibitem[Liu et~al.(2022)Liu, Ning, Cao, Wei, Zhang, Lin, and Hu]{liu2022video}
Ze Liu, Jia Ning, Yue Cao, Yixuan Wei, Zheng Zhang, Stephen Lin, and Han Hu.
\newblock Video swin transformer.
\newblock In \emph{Proceedings of the IEEE/CVF conference on computer vision and pattern recognition}, pages 3202--3211, 2022.

\bibitem[Ma et~al.(2022)Ma, Sun, Rao, Zhou, and Lu]{Ma2022}
Cheng Ma, Haowen Sun, Yongming Rao, Jie Zhou, and Jiwen Lu.
\newblock Video saliency forecasting transformer.
\newblock \emph{IEEE transactions on circuits and systems for video technology}, 32\penalty0 (10):\penalty0 6850--6862, 2022.

\bibitem[Min and Corso(2019)]{min2019tased}
Kyle Min and Jason~J Corso.
\newblock Tased-net: Temporally-aggregating spatial encoder-decoder network for video saliency detection.
\newblock In \emph{Proceedings of the IEEE/CVF International Conference on Computer Vision}, pages 2394--2403, 2019.

\bibitem[Min et~al.(2016)Min, Zhai, Gu, and Yang]{min2016fixation}
Xiongkuo Min, Guangtao Zhai, Ke Gu, and Xiaokang Yang.
\newblock Fixation prediction through multimodal analysis.
\newblock \emph{ACM Transactions on Multimedia Computing, Communications, and Applications (TOMM)}, 13\penalty0 (1):\penalty0 1--23, 2016.

\bibitem[Mishra et~al.(2021)Mishra, Singh, Singh, and Kedia]{mishra2021multi}
Dipti Mishra, Satish~Kumar Singh, Rajat~Kumar Singh, and Divanshu Kedia.
\newblock Multi-scale network (mssg-cnn) for joint image and saliency map learning-based compression.
\newblock \emph{Neurocomputing}, 460:\penalty0 95--105, 2021.

\bibitem[Mital et~al.(2011)Mital, Smith, Hill, and Henderson]{mital2011clustering}
Parag~K Mital, Tim~J Smith, Robin~L Hill, and John~M Henderson.
\newblock Clustering of gaze during dynamic scene viewing is predicted by motion.
\newblock \emph{Cognitive Computation}, 3\penalty0 (1):\penalty0 5--24, 2011.

\bibitem[Moradi et~al.(2024)Moradi, Palazzo, and Spampinato]{moradi2024transformer}
Morteza Moradi, Simone Palazzo, and Concetto Spampinato.
\newblock Transformer-based video saliency prediction with high temporal dimension decoding.
\newblock \emph{arXiv preprint arXiv:2401.07942}, 2024.

\bibitem[Pan et~al.(2017)Pan, Sayrol, Nieto, Ferrer, Torres, McGuinness, and OConnor]{pan2017salgan}
Junting Pan, Elisa Sayrol, Xavier Giro-i Nieto, Cristian~Canton Ferrer, Jordi Torres, Kevin McGuinness, and Noel~E OConnor.
\newblock Salgan: Visual saliency prediction with adversarial networks.
\newblock In \emph{CVPR scene understanding workshop (SUNw)}, 2017.

\bibitem[Peng et~al.(2018)Peng, Zhao, and Zhang]{peng2018two}
Yuxin Peng, Yunzhen Zhao, and Junchao Zhang.
\newblock Two-stream collaborative learning with spatial-temporal attention for video classification.
\newblock \emph{IEEE Transactions on Circuits and Systems for Video Technology}, 29\penalty0 (3):\penalty0 773--786, 2018.

\bibitem[Qiao et~al.(2021)Qiao, Liu, Xu, Deng, Li, Hu, and Borji]{qiao2021joint}
Minglang Qiao, Yufan Liu, Mai Xu, Xin Deng, Bing Li, Weiming Hu, and Ali Borji.
\newblock Joint learning of visual-audio saliency prediction and sound source localization on multi-face videos.
\newblock \emph{arXiv preprint arXiv:2111.08567}, 2021.

\bibitem[Sitzmann et~al.(2020)Sitzmann, Martel, Bergman, Lindell, and Wetzstein]{sitzmann2020implicit}
Vincent Sitzmann, Julien Martel, Alexander Bergman, David Lindell, and Gordon Wetzstein.
\newblock Implicit neural representations with periodic activation functions.
\newblock \emph{Advances in neural information processing systems}, 33:\penalty0 7462--7473, 2020.

\bibitem[Sun et~al.(2017)Sun, Hu, Zhang, Chen, Li, Xie, and Liu]{sun2017}
Xiao Sun, Yuxing Hu, Luming Zhang, Yanxiang Chen, Ping Li, Zhao Xie, and Zhenguang Liu.
\newblock Camera-assisted video saliency prediction and its applications.
\newblock \emph{IEEE transactions on cybernetics}, 48\penalty0 (9):\penalty0 2520--2530, 2017.

\bibitem[Tan et~al.(2024)Tan, Sun, and Liang]{tan2024transformer}
Rui Tan, Minghui Sun, and Yanhua Liang.
\newblock Transformer-based multi-level attention integration network for video saliency prediction.
\newblock \emph{Multimedia Tools and Applications}, pages 1--22, 2024.

\bibitem[Tavakoli et~al.(2019)Tavakoli, Borji, Rahtu, and Kannala]{tavakoli2019dave}
Hamed~R Tavakoli, Ali Borji, Esa Rahtu, and Juho Kannala.
\newblock Dave: A deep audio-visual embedding for dynamic saliency prediction.
\newblock \emph{arXiv preprint arXiv:1905.10693}, 2019.

\bibitem[Tsiami et~al.(2020)Tsiami, Koutras, and Maragos]{tsiami2020stavis}
Antigoni Tsiami, Petros Koutras, and Petros Maragos.
\newblock Stavis: Spatio-temporal audiovisual saliency network.
\newblock In \emph{Proceedings of the IEEE/CVF Conference on Computer Vision and Pattern Recognition}, pages 4766--4776, 2020.

\bibitem[Wang et~al.(2020{\natexlab{a}})Wang, Zhu, Green, Adam, Yuille, and Chen]{wang2020axial}
Huiyu Wang, Yukun Zhu, Bradley Green, Hartwig Adam, Alan Yuille, and Liang-Chieh Chen.
\newblock Axial-deeplab: Stand-alone axial-attention for panoptic segmentation.
\newblock In \emph{European conference on computer vision}, pages 108--126. Springer, 2020{\natexlab{a}}.

\bibitem[Wang et~al.(2020{\natexlab{b}})Wang, Wu, Zhu, Li, Zuo, and Hu]{wang2020eca}
Qilong Wang, Banggu Wu, Pengfei Zhu, Peihua Li, Wangmeng Zuo, and Qinghua Hu.
\newblock Eca-net: Efficient channel attention for deep convolutional neural networks.
\newblock In \emph{Proceedings of the IEEE/CVF conference on computer vision and pattern recognition}, pages 11534--11542, 2020{\natexlab{b}}.

\bibitem[Wang et~al.(2018)Wang, Shen, Guo, Cheng, and Borji]{Wang2018revisiting}
Wenguan Wang, Jianbing Shen, Fang Guo, Ming-Ming Cheng, and Ali Borji.
\newblock Revisiting video saliency: A large-scale benchmark and a new model.
\newblock \emph{Proceedings of the IEEE Conference on Computer Vision and Pattern Recognition (CVPR)}, pages 4894--4903, 2018.

\bibitem[Wang et~al.(2019)Wang, Shen, Xie, Cheng, Ling, and Borji]{wang2019revisiting}
Wenguan Wang, Jianbing Shen, Jianwen Xie, Ming-Ming Cheng, Haibin Ling, and Ali Borji.
\newblock Revisiting video saliency prediction in the deep learning era.
\newblock \emph{IEEE Transactions on Pattern Analysis and Machine Intelligence}, 43\penalty0 (1):\penalty0 220--237, 2019.

\bibitem[Wang et~al.(2021)Wang, Liu, Li, Wang, Zhang, Xu, and Wang]{wang2021spatio}
Ziqiang Wang, Zhi Liu, Gongyang Li, Yang Wang, Tianhong Zhang, Lihua Xu, and Jijun Wang.
\newblock Spatio-temporal self-attention network for video saliency prediction.
\newblock \emph{IEEE Transactions on Multimedia}, 25:\penalty0 1161--1174, 2021.

\bibitem[Wu et~al.(2023)Wu, Zhou, Sun, Gao, Zhu, Zhang, and Yan]{wu2023gfnet}
Songhe Wu, Xiaofei Zhou, Yaoqi Sun, Yuhan Gao, Zunjie Zhu, Jiyong Zhang, and Chenggang Yan.
\newblock Gfnet: gated fusion network for video saliency prediction.
\newblock \emph{Applied Intelligence}, 53\penalty0 (22):\penalty0 27865--27875, 2023.

\bibitem[Wu et~al.(2020)Wu, Wu, Zhang, Ju, and Wang]{Wu2020}
Xinyi Wu, Zhenyao Wu, Jinglin Zhang, Lili Ju, and Song Wang.
\newblock Salsac: A video saliency prediction model with shuffled attentions and correlation-based convlstm.
\newblock In \emph{Proceedings of the AAAI conference on artificial intelligence}, pages 12410--12417, 2020.

\bibitem[Xie et~al.(2024)Xie, Liu, Li, and Song]{xie2024audio}
Jiawei Xie, Zhi Liu, Gongyang Li, and Yingjie Song.
\newblock Audio-visual saliency prediction with multisensory perception and integration.
\newblock \emph{Image and Vision Computing}, 143:\penalty0 104955, 2024.

\bibitem[Xie et~al.(2018)Xie, Sun, Huang, Tu, and Murphy]{xie2018rethinking}
Saining Xie, Chen Sun, Jonathan Huang, Zhuowen Tu, and Kevin Murphy.
\newblock Rethinking spatiotemporal feature learning: Speed-accuracy trade-offs in video classification.
\newblock In \emph{Proc. European Conference on Computer Vision (ECCV)}, pages 305--321, 2018.

\bibitem[Xiong et~al.(2023)Xiong, Wang, Zhang, Huang, Zha, and Zhai]{xiong2023casp}
Junwen Xiong, Ganglai Wang, Peng Zhang, Wei Huang, Yufei Zha, and Guangtao Zhai.
\newblock Casp-net: Rethinking video saliency prediction from an audio-visual consistency perceptual perspective.
\newblock In \emph{Proceedings of the IEEE/CVF Conference on Computer Vision and Pattern Recognition}, pages 6441--6450, 2023.

\bibitem[Xiong et~al.(2024)Xiong, Zhang, You, Li, Huang, and Zha]{xiong2024diffsal}
Junwen Xiong, Peng Zhang, Tao You, Chuanyue Li, Wei Huang, and Yufei Zha.
\newblock Diffsal: joint audio and video learning for diffusion saliency prediction.
\newblock In \emph{Proceedings of the IEEE/CVF Conference on Computer Vision and Pattern Recognition}, pages 27273--27283, 2024.

\bibitem[Yu et~al.(2024)Yu, Sun, Gao, and Gabbouj]{yu2024relevance}
Li Yu, Xuanzhe Sun, Pan Gao, and Moncef Gabbouj.
\newblock Relevance-guided audio visual fusion for video saliency prediction.
\newblock \emph{arXiv preprint arXiv:2411.11454}, 2024.

\bibitem[Zhang and Chen(2018)]{zhang2018video}
Kao Zhang and Zhenzhong Chen.
\newblock Video saliency prediction based on spatial-temporal two-stream network.
\newblock \emph{IEEE Transactions on Circuits and Systems for Video Technology}, 29\penalty0 (12):\penalty0 3544--3557, 2018.

\bibitem[Zhang et~al.(2023)Zhang, Zhang, Wu, and Tao]{zhang2023multi}
Yunzuo Zhang, Tian Zhang, Cunyu Wu, and Ran Tao.
\newblock Multi-scale spatiotemporal feature fusion network for video saliency prediction.
\newblock \emph{IEEE Transactions on Multimedia}, 26:\penalty0 4183--4193, 2023.

\bibitem[Zhou et~al.(2023)Zhou, Wu, Shi, Zheng, Wang, Yin, Zhang, and Yan]{zhou2023transformer}
Xiaofei Zhou, Songhe Wu, Ran Shi, Bolun Zheng, Shuai Wang, Haibing Yin, Jiyong Zhang, and Chenggang Yan.
\newblock Transformer-based multi-scale feature integration network for video saliency prediction.
\newblock \emph{IEEE Transactions on Circuits and Systems for Video Technology}, 33\penalty0 (12):\penalty0 7696--7707, 2023.

\bibitem[Zhou et~al.(2021)Zhou, Pei, Li, Wang, Zheng, and He]{zhou2021saliency}
Zikun Zhou, Wenjie Pei, Xin Li, Hongpeng Wang, Feng Zheng, and Zhenyu He.
\newblock Saliency-associated object tracking.
\newblock In \emph{Proceedings of the IEEE/CVF international conference on computer vision}, pages 9866--9875, 2021.

\bibitem[Zhu et~al.(2024{\natexlab{a}})Zhu, Zhang, Zhu, Zhang, Ding, Zhai, and Yang]{zhu2024discrete}
Dandan Zhu, Kaiwei Zhang, Kun Zhu, Nana Zhang, Weiping Ding, Guangtao Zhai, and Xiaokang Yang.
\newblock From discrete representation to continuous modeling: A novel audio-visual saliency prediction model with implicit neural representations.
\newblock \emph{IEEE Transactions on Emerging Topics in Computational Intelligence}, 2024{\natexlab{a}}.

\bibitem[Zhu et~al.(2024{\natexlab{b}})Zhu, Duan, Zhang, Zhu, Zhu, Teng, Min, and Zhai]{zhu2024does}
Yuxin Zhu, Huiyu Duan, Kaiwei Zhang, Yucheng Zhu, Xilei Zhu, Long Teng, Xiongkuo Min, and Guangtao Zhai.
\newblock How does audio influence visual attention in omnidirectional videos? database and model.
\newblock \emph{arXiv preprint arXiv:2408.05411}, 2024{\natexlab{b}}.

\end{thebibliography}
}

\clearpage
\setcounter{page}{1}
\maketitlesupplementary

\appendix

\section{Dataset Details}
\noindent \textbf{Visual Datasets:} DHF1K \cite{Wang2018revisiting} is a large-scale benchmark dataset for video saliency prediction. It consists of 1,000 videos, with 600 training, 100 validation, and 300 test videos. 
The UCF-Sports \cite{peng2018two} is a task-driven dataset sourced from broadcast TV channels. The dataset is divided into 103 training videos and 47 testing videos~\cite{wang2021spatio}.


\noindent \textbf{Audio-Visual Datasets:} Six audio-visual datasets are utilized for our evaluation: DIEM \cite{mital2011clustering}, ETMD \cite{koutras2015perceptually}, SumMe \cite{gygli2014creating}, AVAD \cite{min2016fixation}, Coutrot1 \cite{coutrot2014saliency}, and Coutrot2 \cite{coutrot2016multimodal}. DIEM contains 84 video clips covering various content types such as commercials, documentaries, sports events, and movie trailers. ETMD comprises 12 clips extracted from diverse Hollywood films. SumMe consists of 25 unstructured video clips captured in a controlled experiment, showcasing themes ranging from sports and cooking to travel. AVAD includes 45 short clips featuring a variety of audio-visual scenarios, such as musical performances and sports activities. Coutrot1 offers 60 clips divided into four visual categories—individual/group dynamics, natural landscapes, multiple moving objects, and close-ups of faces—with eye-tracking data collected from 72 participants. In contrast, Coutrot2 comprises 15 clips capturing a meeting, accompanied by eye-tracking data from 40 observers. ~\autoref{tab:detail-dataset} presents a list of these datasets along with
specific details about each.

\begin{table}[htbp]
    \centering
    \caption{Information on Video Saliency Datasets} \label{tab4}
	\resizebox{\linewidth}{!}{%
    \begin{tabular}{lcccccc}
        \toprule
        Dataset & Year & Videos & Resolution & Viewers & Frames & Audio \\
        \midrule
        DH1FK & 2017 & 1,000 & 360 × 640 & 17 & 582,605 & No \\
        UCF-Sports & 2012 & 150 & 480 × 720 & 12 & 9,580 & No \\
        DIEM & 2011 & 84 & 720 × 1280 & 42 & 240,452 & Yes \\
        AVAD & 2015 & 45 & 720 × 1280 & 16 & 9,564 & Yes \\
        Coutrot-1 & 2013 & 60 & 576 × 720 & 72 & 25,223 & Yes \\
        Coutrot-2 & 2014 & 15 & 576 × 720 & 40 & 17,134 & Yes \\
        SumMe & 2014 & 25 & 256 × 340 & 10 & 109,788 & Yes \\
        ETMD & 2015 & 12 & 280 × 356 & 10 & 52,744 & Yes \\
        \bottomrule
    \end{tabular}
    }
    \label{tab:detail-dataset}
\end{table}

\begin{table*}[htbp]
\centering
\scriptsize
\caption{Comparison with Previous Methods on DHF1K and UCF Sports Datasets. For our model, we indicate the percentage (\%) change in performance relative to the second-best result, or to the best result if ours is not the top performer. The best results are highlighted in \textcolor{red}{red}, the second-best in \textcolor{blue}{blue}, and the third-best in \textcolor{darkgreen}{green}.}
\label{tab:comparison}
\resizebox{1\linewidth}{!}{%
\begin{tabular}{c|cccc|cccc}
\hline
			\textbf{Model} & \multicolumn{4}{c|}{\textbf{DHF1K}} & \multicolumn{4}{c}{\textbf{UCF Sports}} \\
			\cmidrule(lr){2-5} \cmidrule(lr){6-9}
			& \textbf{CC $\uparrow$} & \textbf{NSS $\uparrow$} & \textbf{SIM $\uparrow$} & \textbf{AUC-J $\uparrow$} & \textbf{CC $\uparrow$} & \textbf{NSS $\uparrow$} & \textbf{SIM $\uparrow$} & \textbf{AUC-J $\uparrow$} \\
			\hline
			DeepVS \cite{jiang2018deepvs} & $0.344$ & $1.911$ & $0.256$ & $0.856$ & $0.405$ & $2.089$ & $0.321$ & $0.870$ \\
			ACLNet \cite{wang2019revisiting} & $0.434$ & $2.354$ & $0.315$ & $0.890$ & $0.510$ & $2.567$ & $0.406$ & $0.897$ \\
			STRA-Net  \cite{lai2019video} & $0.458$ & $2.558$ & $0.355$ & $0.895$ & $0.593$ & $3.018$ & $0.479$ & $0.910$ \\
			TASED-Net \cite{min2019tased} & $0.470$ & $2.667$ & $0.361$ & $0.895$ & $0.582$ & $2.920$ & $0.469$ & $0.899$ \\
			SalSAC \cite{Wu2020} & $0.479$ & $2.673$ & $0.357$ & $0.896$ & $0.671$ & $3.523$ & $0.534$ & $0.926$ \\
			UNISAL \cite{droste2020unified} & $0.490$ & $2.776$ & $0.390$ & $0.901$ & $0.644$ & $3.381$ & $0.523$ & $0.918$ \\
			ECANet \cite{wang2020eca} & $0.500$ & $2.814$ & $0.385$ & $0.903$ & $0.673$ & $3.620$ & $0.522$ & $0.924$ \\
			HD2S \cite{bellitto2021hierarchical} & $0.503$ & $2.812$ & $0.406$ & $0.908$ & $0.604$ & $3.114$ & $0.507$ & $0.904$ \\
			ViNet \cite{jain2021vinet} & $0.511$ & $2.872$ & $0.381$ & $0.908$ & $0.673$ & $3.620$ & $0.522$ & $0.924$ \\
			TSFP-Net \cite{chang2021temporal} & $0.517$ & $2.967$ & $0.392$ & $0.912$ & $0.685$ & $3.698$ & $0.561$ & $0.923$ \\
			GFNet  \cite{wu2023gfnet} & $0.526$ & $2.995$ & $0.379$ & $0.913$ & $0.694$ & $3.723$ & $0.544$ & $0.933$ \\
			STSANet \cite{wang2021spatio} & $0.529$ & $3.010$ & $0.383$ & $0.913$ & $0.705$ & $3.908$ & $0.560$ & $0.936$ \\
			TMFI-Net  \cite{zhou2023transformer} & $0.546$ & \textcolor{darkgreen}{$3.146$} & \textcolor{darkgreen}{$0.407$} & \textcolor{darkgreen}{$0.9153$} & $0.707$ & $3.863$ & $0.565$ & $0.936$ \\
            TinyHD-S \cite{hu2023tinyhd} & $0.492$ & $2.873$ & $0.388$ & $0.907$ & $0.624$ & $ 3.280$ & $0.510$ & $0.918$ \\
		THTD-Net \cite{moradi2024transformer} & \textcolor{darkgreen}{$0.547$} & $3.138$ & $0.406$ & $0.9152$ & $0.711$ & $3.840$ & $0.565$ & $0.933$ \\
			TempVST \cite{lazaridis2024visual} & $0.494$ & $3.73$ & $0.373$ & $0.904$ & $-$ & $-$ & $-$ & $-$ \\
            ViNet-E \cite{girmaji2025minimalistic} & $0.549$ & $3.13$ & $0.409$ & $0.922$ & \textcolor{blue}{$0.744$} & \textcolor{red}{$4.156$} & \textcolor{blue}{$0.587$} & \textcolor{blue}{$0.941$} \\
			DiffSal(V) \cite{xiong2024diffsal} & $0.533$ & $3.066$ & $0.405$ & \textcolor{blue}{$0.918$} & $0.685$ & $3.483$ & $0.543$ & $0.928$ \\
			HSFI-Net \cite{jin2025hierarchical} & \textcolor{blue}{$0.548$} & \textcolor{blue}{$3.163$} & \textcolor{blue}{$0.411$} & $0.914$ & \textcolor{darkgreen}{$0.714$} & \textcolor{darkgreen}{$3.896$} & \textcolor{darkgreen}{$0.568$} & \textcolor{darkgreen}{$0.937$} \\
			\midrule
			\rowcolor[HTML]{DADCFF} \textbf{DTFSal(V)} 
			& \textcolor{red}{$0.561$\textsubscript{\scriptsize{+2.37\%}}} 
			& \textcolor{red}{$3.205$\textsubscript{\scriptsize{+1.33\%}}} 
			& \textcolor{red}{$0.436$\textsubscript{\scriptsize{+6.08\%}}} 
			& \textcolor{red}{$0.923$\textsubscript{\scriptsize{+0.55\%}}}
			& \textcolor{red}{$0.768$\textsubscript{\scriptsize{+3.76\%}}} 
			& \textcolor{blue}{$4.04$\textsubscript{\scriptsize{-2.78\%}}} 
			& \textcolor{red}{$0.614$\textsubscript{\scriptsize{+4.60\%}}} 
			& \textcolor{red}{$0.947$\textsubscript{\scriptsize{+0.64\%}}}\\       
			\hline
	\end{tabular}}
\end{table*}

\vspace{-3pt}

\section{Supplementary Experiments}

\vspace{-3pt}

\noindent \textbf{Comparison with Video Saliency prediction methods:} ~\autoref{tab:comparison} presents a detailed comparison of our visual-only variant, DTFSal(V), against recent SOTA video saliency prediction methods on DHF1K and UCF Sports datasets. Our results indicate that DTFSal(V) achieves the best performance in 7 evaluation metrics and ranks second in the remaining one. Notably, it outperforms the second-best method by a relative margin of 2.06\% across these datasets. These results further highlight the strong performance of our approach, demonstrating that even the visual-only variant can achieve competitive SOTA results when compared to methods such as ViNet-E~\cite{girmaji2025minimalistic} and HSFI-Net~\cite{jin2025hierarchical}. These findings clearly demonstrate the effectiveness of our approach for visual saliency prediction.

\vspace{3pt}

\noindent \textbf{Role of Audio Information:} To comprehensively analyze the impact of audio information on audio-visual saliency prediction, we conduct experiment on multiple benchmark datasets. Specifically, we evaluate our model's performance by removing the audio feature extraction branch while keeping all other settings unchanged. The results, as presented in ~\autoref{tab:table-av-sota} in the \textcolor{blue}{main paper} and~\autoref{tab:table-v-sota}, clearly demonstrate that incorporating audio cues significantly enhances saliency prediction performance across all six datasets. This improvement is primarily attributed to the ability of audio features to help localize sound-emitting objects more effectively, thereby capturing crucial cross-modal correspondences between audio and visual stimuli \cite{zhu2024does, zhu2024discrete}. Notably, our proposed DTFSal(V) model achieves competitive or superior performance compared to SOTA methods, demonstrating its effectiveness in leveraging auditory cues for improved saliency prediction. The consistent performance boost highlights the importance of integrating audio information to refine attention mechanisms and enhance prediction accuracy.

\begin{table*}[!tbp]
	\begin{center}
		\caption{Comparison of different visual models on six audio-visual benchmark datasets. In this evaluation, we remove the audio feature extraction component and assess the performance of visual-only models on audio-visual datasets. The best results are highlighted in \textcolor{red}{red} and the second-best in \textcolor{blue}{blue}.}
		\label{tab:table-v-sota}
		\vspace*{-5pt}
		\resizebox{1\linewidth}{!}{
			\begin{tabular}{c | c | cccc | cccc | cccc}
				\toprule
				\textbf{Method} & \textbf{\#Parameters} & \multicolumn{4}{c|}{\textbf{DIEM}} & \multicolumn{4}{c|}{\textbf{Coutrot1}} & \multicolumn{4}{c}{\textbf{Coutrot2}}\\
				\cline{3-14}
				& & CC $\uparrow$ & NSS $\uparrow$ & AUC-J $\uparrow$  & SIM  $\uparrow$ & CC $\uparrow$ & NSS $\uparrow$ & AUC-J $\uparrow$   & SIM $\uparrow$ & CC $\uparrow$ & NSS $\uparrow$ & AUC-J $\uparrow$  & SIM $\uparrow$ \\
				\hline
				ACLNet & - & $0.522$ & $2.02$ & $0.869$ & $0.427$ & $0.425$ & $1.92$ & $0.850$ & $0.361$ & $0.448$ & $3.16$ & $0.926$ & $0.322$ \\
				TASED-Net & $21.26M$ & $0.557$ & $2.16$ & $0.881$ & $0.461$ & $0.479$ & $2.18$ & $0.867$ & $0.388$ & $0.437$ & $3.17$ & $0.921$ & $0.314$ \\
				STAViS & $20.76M$ & $0.579$ & $2.26$ & $0.883$ & $0.482$ & $0.472$ & $2.11$ & $0.868$ & $0.393$ & $0.734$ & $5.28$ & $0.958$ & $0.511$ \\
				ViNet & $33.97M$ & $0.632$ & $2.53$ & $0.899$ & $0.498$ & $0.560$ & $2.73$ & $0.889$ & $0.425$ & $0.754$ & $5.95$ & $0.951$ & $0.493$ \\
				TSFP-Net & - & $0.651$ & $2.62$ & $0.906$ & $0.527$ & $0.571$ & $2.73$ & $0.895$ & $0.447$ & $0.743$ & $5.31$ & $0.959$ & $0.528$ \\
				CASP-Net(V) & $51.62M$ & $0.649$ & $2.59$ & $0.904$ & $0.538$ & $0.559$ & $2.64$ & $0.888$ & $0.445$ & $0.756$ & $6.07$ & $0.963$ & $0.567$ \\ 
                RAVF & - & $0.668$ & $2.67$ & $0.907$ & $0.545$ & $0.590$ & $2.83$ & $0.895$ & $0.459$ & $0.820$ & $6.28$ & $0.963$ & $0.602$ \\
                TM2SP-Net & - & $0.676$ & $2.61$ & $0.909$ & $0.539$ & \textcolor{blue}{$0.615$} & \textcolor{blue}{$3.16$} & $0.903$ & \textcolor{blue}{$0.492$} & $0.786$ & $6.13$ & \textcolor{blue}{$0.963$} & $0.598$ \\ 
				ViNet-E & $48.19M$ & \textcolor{red}{$0.701$} & \textcolor{red}{$2.84$} & \textcolor{red}{$0.913$} & \textcolor{red}{$0.566$} & $0.614$ & $3.08$ & \textcolor{red}{$0.905$} & $0.465$ & \textcolor{red}{$0.854$} & \textcolor{red}{$6.76$} & $0.962$ & \textcolor{red}{$0.628$} \\
                \midrule
				\rowcolor[HTML]{DADCFF} \textbf{DTFSal(V)} & $40.73M$ & \textcolor{blue}{$0.677$} & \textcolor{blue}{$2.76$} & \textcolor{blue}{$0.910$} & \textcolor{blue}{$0.548$} & \textcolor{red}{$0.623$} & \textcolor{red}{$3.21$} & \textcolor{blue}{$0.904$} & \textcolor{red}{$0.505$} & \textcolor{blue}{$0.842$} & \textcolor{blue}{$6.42$} & \textcolor{red}{$0.965$} & \textcolor{blue}{$0.618$} \\
				\hline
				\multirow{2}*{\textbf{Method}}  & \multirow{2}*{\textbf{\#Parameters}} & \multicolumn{4}{c|}{\textbf{AVAD}} & \multicolumn{4}{c|}{\textbf{ETMD}} & \multicolumn{4}{c}{\textbf{SumMe}}\\
				\cline{3-14}
				& & CC $\uparrow$ & NSS $\uparrow$ & AUC-J $\uparrow$  & SIM  $\uparrow$ & CC $\uparrow$ & NSS $\uparrow$ & AUC-J $\uparrow$   & SIM $\uparrow$ & CC $\uparrow$ & NSS $\uparrow$ & AUC-J $\uparrow$  & SIM $\uparrow$ \\
				\hline
				ACLNet & - & $0.580$ & $3.17$ & $0.905$ & $0.446$ & $0.477$ & $2.36$ & $0.915$ & $0.329$ & $0.379$ & $1.79$ & $0.868$ & $0.296$ \\
				TASED-Net & $21.26M$ & $0.601$ & $3.16$ & $0.914$ & $0.439$ & $0.509$ & $2.63$ & $0.916$ & $0.366$ & $0.428$ & $2.10$ & $0.884$ & $0.333$ \\
				STAViS & $20.76M$ & $0.608$ & $3.18$ & $0.919$ & $0.457$ & $0.569$ & $2.94$ & $0.931$ & $0.425$ & $0.422$ & $2.04$ & $0.888$ & $0.337$ \\
				ViNet & $33.97M$ & $0.674$ & $3.77$ & $0.927$ & $0.491$ & $0.571$ & $3.08$ & $0.928$ & $0.406$ & $0.463$ & $2.41$ & $0.897$ & $0.343$ \\
				TSFP-Net & - & $0.704$ & $3.77$ & $0.932$ & $0.521$ & $0.576$ & $3.07$ & $0.932$ & $0.428$ & $0.464$ & $2.30$ & $0.894$ & $0.360$ \\
				CASP-Net(V) & $51.62M$ & $0.681$ & $3.75$ & $0.931$ & $0.526$ & $0.616$ & $3.31$ & \textcolor{red}{$0.938$} & \textcolor{blue}{$0.471$} & $0.485$ & $2.52$ & $0.904$ & $0.382$ \\
                RAVF(V) & - & $0.697$ & $3.75$ & $0.933$ & $0.533$ & $0.611$ & $3.33$ & $0.930$ & $0.470$ & $0.494$ & \textcolor{blue}{$2.57$} & $0.900$ & $0.374$ \\
                TM2SP-Net & - & \textcolor{blue}{$0.730$} & $4.12$ & \textcolor{blue}{$0.934$} & \textcolor{red}{$0.568$} & $0.616$ & $3.33$ & $0.940$ & $0.470$ & \textcolor{blue}{$0.505$} & $2.53$ & \textcolor{blue}{$0.906$} & \textcolor{blue}{$0.388$} \\ 
				ViNet-E & $48.19M$ & $0.729$ & \textcolor{blue}{$4.16$} & $0.938$ & $0.547$ & \textcolor{blue}{$0.632$} & $3.43$ & $0.943$ & $0.468$ & - & - & - & - \\
				\midrule
				\rowcolor[HTML]{DADCFF} \textbf{DTFSal(V)} & $40.73M$ & \textcolor{red}{$0.742$} & \textcolor{red}{$4.21$} & \textcolor{red}{$0.938$} & \textcolor{blue}{$0.559$} & \textcolor{red}{$0.653$} & \textcolor{red}{$3.57$} & \textcolor{blue}{$0.939$} & \textcolor{red}{$0.499$} & \textcolor{red}{$0.561$} & \textcolor{red}{$2.91$} & \textcolor{red}{$0.911$} & \textcolor{red}{$0.441$} \\
				\bottomrule
		\end{tabular}
		}
	\end{center}
	\vspace{-10pt}
\end{table*}

\begin{figure*}[t]
    \centering
    \includegraphics[width=\textwidth]{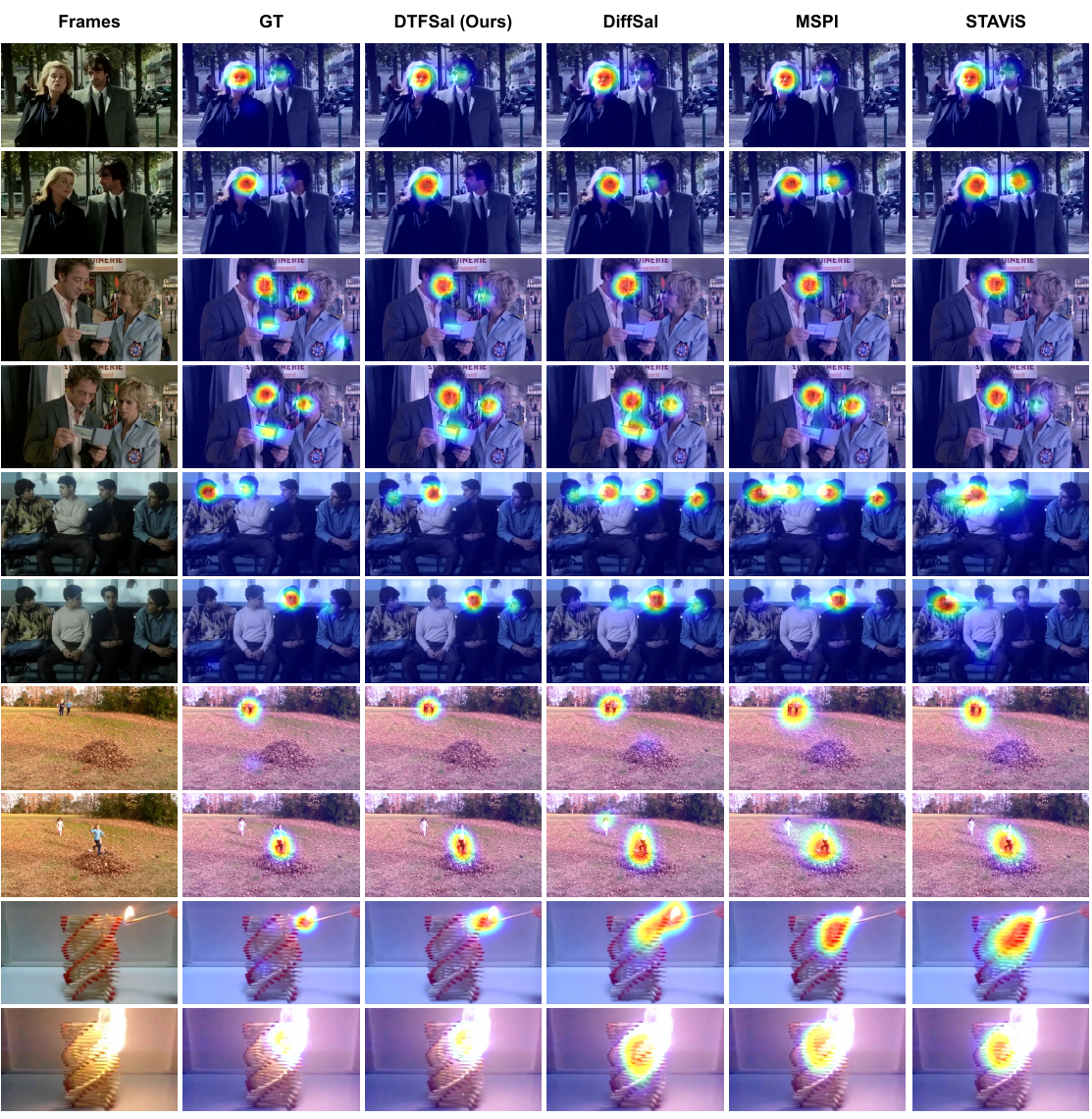}
    \caption{Additional comparative visualizations of our DTFSal model compared with previous SOTA audio-visual saliency prediction methods.}
    \label{fig:qualitative_result_fig_AV_supp}
\end{figure*}

\begin{figure*}[t]
    \centering
    \includegraphics[width=\textwidth]{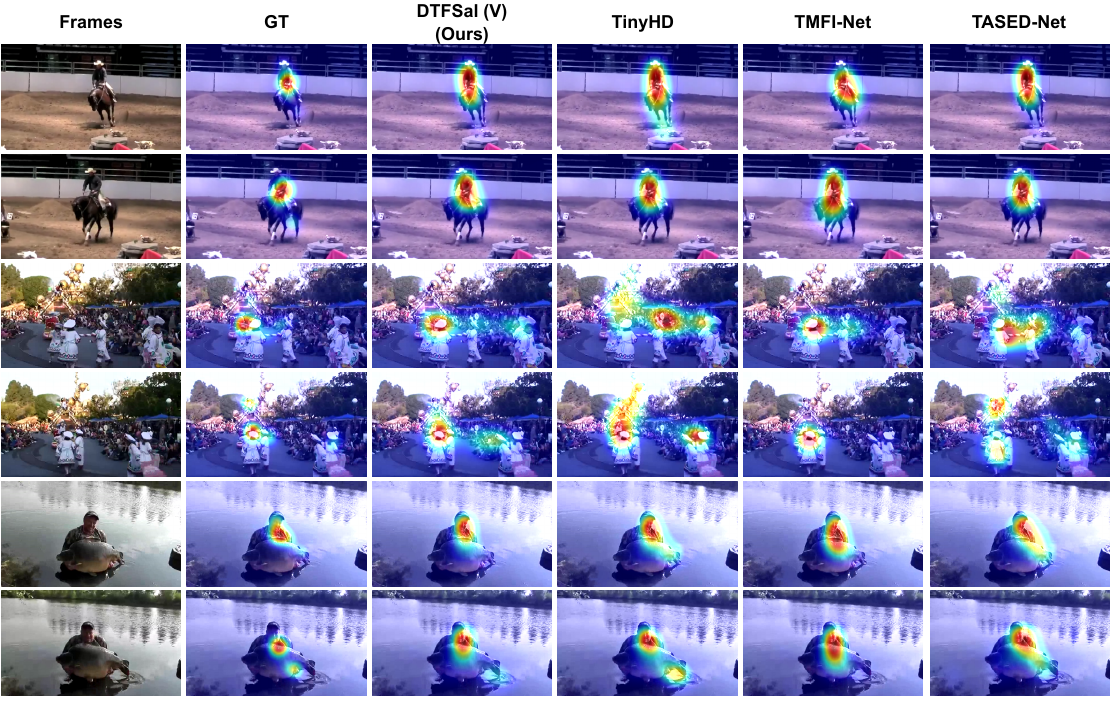}
    \caption{Comparative visualizations of our DTFSal model compared with previous SOTA visual-only saliency prediction methods.}
    \label{fig:qualitative_result_fig_V_supp}
\end{figure*}

\noindent \textbf{More Qualitative Results:}~\autoref{fig:qualitative_result_fig_AV_supp} presents additional qualitative results comparing our DFTSal model with recent audio-visual saliency prediction methods on various audio-visual datasets. As shown, the predictions of our model align more closely with the ground truth (GT), highlighting its effectiveness.
Furthermore,~\autoref{fig:qualitative_result_fig_V_supp} shows the visualization results of our visual-only variant, DFTSal(V), against the SOTA visual saliency models. The results demonstrate that DFTSal(V) generates saliency maps that closely match the ground truth, effectively capturing key visual attention cues.

\vspace{3pt}


\vspace{-3pt}

\section{Supplementary Ablation Studies}

\vspace{-3pt}

\noindent \textbf{Analyzing the Performance of Model Using Different Video Encoders:} We conduct experiments on DTFSal using the video encoders employed in other SOTA models~\cite{tsiami2020stavis, zhou2023transformer}. ~\autoref{tab:video_encoders_comparison} compares the performance of our model on DHF1K dataset with these encoders. While S3D~\cite{xie2018rethinking} and Video Swin-S~\cite{liu2022video} deliver competitive results, MViTv2 ~\cite{li2022mvitv2} achieves the highest scores across all metrics with a moderate parameter count, demonstrating its effectiveness in capturing spatio-temporal features for saliency prediction.

\begin{table}[ht]
    \centering
    \caption{Impact of Using Different Video Encoders.}
    \label{tab:video_encoders_comparison}
    \resizebox{\linewidth}{!}{
    \begin{tabular}{lccccc}
        \hline
        Video Encoder & CC $\uparrow$ & NSS $\uparrow$ & AUC-J $\uparrow$ & SIM $\uparrow$ & Number of Parameters (Million) \\ \hline
        S3D~\cite{xie2018rethinking}  & 0.543 & 2.954 & 0.914 & 0.396 & 8.77 M \\
        Video Swin-S~\cite{liu2022video} & 0.542 & 3.08 & 0.920 & 0.419 & 49.8 M \\
         \rowcolor[HTML]{DADCFF} MViTv2~\cite{li2022mvitv2} & \textbf{0.561} & \textbf{3.205} & \textbf{0.923} & \textbf{0.446} & 35 M \\
        \hline
    \end{tabular}}
\end{table}

\vspace{3pt}

\noindent \textbf{Impact of Pre-trained Encoder Weights:} Initializing the visual branch with pre-trained encoder weights from action recognition datasets such as Kinetics-400~\cite{carreira2017quo} significantly boosts performance on DHF1K compared to training from scratch. Using these pre-trained weights provides a strong feature representation, leading to notable improvements across all performance metrics.~\autoref{tab:pretrained_weights} highlights this impact, demonstrating that leveraging prior knowledge from large-scale datasets enhances the model’s ability to capture spatiotemporal patterns, making it more effective for saliency prediction \cite{pan2017salgan,min2019tased}.

\begin{table}[h]
    \centering
    \caption{Impact of Pre-trained Encoder Weights on Model Performance for DHF1K dataset. The best results are marked in bold.}
    \label{tab:pretrained_weights}
    \resizebox{\linewidth}{!}{%
    \begin{tabular}{lcccc}
        \hline
        \textbf{Weights Usage} & \textbf{CC $\uparrow$} & \textbf{NSS $\uparrow$} & \textbf{AUC-J $\uparrow$} & \textbf{SIM $\uparrow$} \\ \hline
        Without Pre-trained Weights & $0.400$ & $2.149$ & $0.879$ & $0.324$ \\
        \rowcolor[HTML]{DADCFF} With Pre-trained Weights & $\textbf{0.561}$ & $\textbf{3.205}$ & $\textbf{0.923}$ & $\textbf{0.446}$ \\ \hline
    \end{tabular}}
\end{table}

\vspace{3pt}

\noindent \textbf{Impact of Pre-trained DHF1K Weights on Audio-Visual Model Performance:} We investigate the impact of different training strategies, as shown in~\autoref{tab:ablation_freeze}. The three strategies include training without pre-trained weights on DHF1K, freezing the DHF1K-trained model and training only the audio fusion module, and fine-tuning the full model after initializing with pretrained weights on DHF1K. The results show that fine-tuning the full model achieves the best performance, as it enables the model to learn rich visual saliency features and better integrate audio-visual cues. Moreover, the findings confirm that pre-training on a large-scale video dataset enhances generalization to other datasets, which is particularly beneficial given the limited size of audio-visual datasets.

\begin{table}[!]
\centering
\caption{Impact of different training strategies on ETMD and AVAD datasets. The best results are marked in bold.}
\label{tab:ablation_freeze}
\resizebox{\linewidth}{!}{%
\begin{tabular}{l|cc|cc}
\toprule
\multirow{2}{*}{Training Strategy} & \multicolumn{2}{c|}{ETMD} & \multicolumn{2}{c}{AVAD} \\
\cmidrule(lr){2-3} \cmidrule(lr){4-5}
 & CC$\uparrow$ & SIM$\uparrow$ & CC$\uparrow$ & SIM$\uparrow$ \\
\midrule
Without pre-trained DHF1K weights     & 0.628 & 0.479 & 0.682 & 0.525 \\
Freeze pre-trained weights, Train only Audio Fusion  & 0.642 & 0.482 & 0.700 & 0.532 \\
\rowcolor[HTML]{DADCFF} Fine-tune Full Model    & \textbf{0.667} & \textbf{0.531} & \textbf{0.746} & \textbf{0.595} \\
\bottomrule
\end{tabular}%
}
\end{table}

\vspace{3pt}

\noindent \textbf{Impact of Video Clip Length on Model Performance:} We analyze the impact of different video clip lengths ($T$) on model performance, as shown in~\autoref{tab:comparison_tol_video}. This experiment is conducted on DHF1K dataset. The results show that a clip length of $32$ frames achieves the best balance for temporal fitting during training. Shorter clips may not capture sufficient temporal information, leading to suboptimal performance. Conversely, longer clips, while potentially offering slight improvements, increase computational complexity and may lead to overfitting.

\begin{table}[!h]
	\centering
	\caption{Quantitative Comparison of the Model Using Video Clips of Different Lengths on DHF1K dataset.}
	\label{tab:comparison_tol_video}
        \resizebox{\linewidth}{!}{%
	\begin{tabular}{ccccc}
		\hline
		\textbf{Clip Length $(T)$} & \textbf{$\text{AUC-J} \uparrow$} & \textbf{$\text{SIM} \uparrow$} & \textbf{$\text{CC} \uparrow$} & \textbf{$\text{NSS} \uparrow$} \\ 
		\hline
		$4$  & $0.908$ & $0.409$ & $0.506$ & $2.848$ \\ 
		$8$  & $0.910$ & $0.419$ & $0.512$ & $2.513$ \\ 
		$16$ & $0.917$ & $0.434$ & $0.547$ & $3.044$ \\ 
		  \rowcolor[HTML]{DADCFF} $32$ & $\textbf{0.923}$ & $\textbf{0.441}$ & $\textbf{0.561}$ & $\textbf{3.205}$ \\ 
		$48$ & $0.920$ & $0.436$ & $0.557$ & $3.166$ \\ 
		\hline
	\end{tabular}}
\end{table}

\begin{table}[!tbp]
\centering
\caption{Impact of different training losses on DHF1K. The best results are marked in bold.}
\label{tab:loss-ablation}
\resizebox{0.8\linewidth}{!}{%
\begin{tabular}{l|cc|ccc}
\toprule
\multirow{2}{*}{Method} & \multicolumn{2}{c|}{Losses} & \multicolumn{3}{c}{DHF1K} \\
\cmidrule(lr){2-3}\cmidrule(lr){4-6}
 & $\mathcal{L}_{KL}$ & $\mathcal{L}_{CC}$ & CC$\uparrow$ & NSS$\uparrow$ & SIM$\uparrow$ \\
\midrule
I          & \checkmark &  \ding{55}  & 0.558 & 3.173 & 0.434 \\
II         &    \ding{55}       & \checkmark & 0.557 & 3.177 & 0.433 \\
\rowcolor[HTML]{DADCFF} III(Ours)  &  \checkmark & \checkmark & \textbf{0.561} & \textbf{3.205} & \textbf{0.436} \\
\bottomrule
\end{tabular}%
}
\end{table}

\begin{table}[!tbp]
\centering
\caption{Impact of different training losses on ETMD and AVAD. The best results are marked in bold.}
\label{tab:loss-ablation_AV}
\resizebox{\linewidth}{!}{%
\begin{tabular}{l|cc|ccc|ccc}
\toprule
\multirow{2}{*}{Method} & \multicolumn{2}{c|}{Losses} 
                        & \multicolumn{3}{c|}{ETMD} 
                        & \multicolumn{3}{c}{AVAD} \\
\cmidrule(lr){2-3}\cmidrule(lr){4-6}\cmidrule(lr){7-9}
 & $\mathcal{L}_{KL}$ & $\mathcal{L}_{CC}$ 
 & CC$\uparrow$ & NSS$\uparrow$ & SIM$\uparrow$ 
 & CC$\uparrow$ & NSS$\uparrow$ & SIM$\uparrow$ \\
\midrule
I          
& \checkmark &   \ding{55}         
& 0.668 & 3.67 & 0.646
& 0.684 & 3.75 & 0.528 \\

II         
&    \ding{55}       & \checkmark 
& 0.606 & 3.26 & 0.460 
& 0.696 & 3.78 & 0.540 \\

\rowcolor[HTML]{DADCFF} III (Ours)  
& \checkmark & \checkmark 
& \textbf{0.667} & \textbf{3.82} & \textbf{0.531}
& \textbf{0.756} & \textbf{4.36} & \textbf{0.595} \\
\bottomrule
\end{tabular}%
}
\end{table}

\vspace{5pt}

\begin{figure*}[!htbp]
    \centering
    \includegraphics[width=\textwidth]{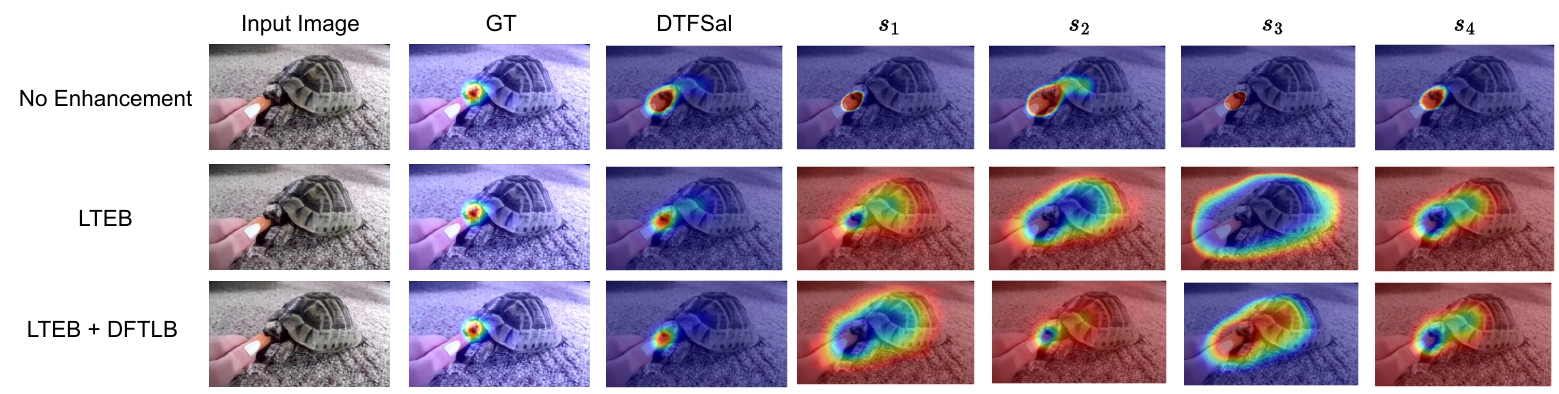}
    \caption{Decoder-based saliency predictions (\(s_1\)--\(s_4\), final prediction) showing the impact of LTEB and DLTFB compared to baseline.}
    \label{fig:decoder_supp}
\end{figure*}
\vspace{3pt}

\noindent \textbf{Impact of different training losses:} We evaluate the effect of different loss combinations by training our model using $\mathcal{L}_{KL}$ and $\mathcal{L}_{CC}$ individually and in combination. As shown in ~\autoref{tab:loss-ablation} for DHF1K, using $\mathcal{L}_{KL}$ alone results in better performance than using $\mathcal{L}_{CC}$ alone, yet the best performance is achieved when both losses are combined. Utilizing a composite loss function that blends multiple evaluation metrics leads to a more efficient training process and ultimately improves performance~\cite{bruckert2021deep}. Similar trends are observed in the audio-visual evaluation on ETMD \cite{koutras2015perceptually} and AVAD \cite{min2016fixation} (see~\autoref{tab:loss-ablation_AV}), confirming that incorporating both losses leads to more effective saliency prediction, with the KL component playing a more dominant role.

\vspace{3pt}

\noindent \textbf{Visualization of Saliency Predictions from Decoders}
To provide a more intuitive understanding of the contributions of our multi-decoder, we visualized the saliency map predictions from each decoder (\(S_1\), \(S_2\), \(S_3\), \(S_4\)) as well as the final saliency map \(S\) under different configurations: a baseline model without any enhancement modules, a model with only LTEB, and the full model incorporating both LTEB and DLTFB. As shown in~\autoref{fig:decoder_supp}, the incorporation of LTEB and DLTFB progressively enhances the feature representations and refines the predicted saliency, leading to more accurate and context-aware results compared to the ground truth.

\section{Limitation \& Future works}
A key limitation of our work is the lack of large-scale audio-visual datasets, which constrains broader generalization. Future efforts could focus on collecting more comprehensive datasets across diverse domains—such as commercials, documentaries, and other complex scenes—and on unifying image-level and video-level human visual modeling under a single framework.

\end{document}